\documentclass{article}

\usepackage[nonatbib,preprint]{neurips_2020}

\usepackage[utf8]{inputenc} 
\usepackage[T1]{fontenc}    
\usepackage{hyperref}       
\usepackage{url}            
\usepackage{booktabs}       
\usepackage{amsfonts}       
\usepackage{nicefrac}       
\usepackage{microtype}      

\usepackage{graphicx}
\usepackage{caption}
\usepackage{subcaption}
\usepackage{float}
\usepackage{diagbox}

\usepackage{booktabs}
\usepackage{placeins}

\title{Can Reinforcement Learning for Continuous Control Generalize Across Physics Engines?}

\author{
  Aaqib Parvez Mohammed\\
  Department of Computer Science\\
  Hochschule Bonn-Rhein-Sieg\\
  Sankt Augustin, Germany\\
  \texttt{aaqib49@gmail.com} \\
  \And
  Matias Valdenegro-Toro \\
  Robotics Innovation Center\\
  German Research Center for Artificial Intelligence\\
  Bremen, Germany \\
  \texttt{matias.valdenegro@dfki.de} \\
}

\begin{document}
\maketitle


\begin{abstract}
    Reinforcement learning (RL) algorithms should learn as much as possible about the environment but not the properties of the physics engines that generate the environment. There are multiple algorithms that solve the task in a physics engine based environment but there is no work done so far to understand if the RL algorithms can generalize across physics engines. In this work, we compare the generalization performance of various deep reinforcement learning algorithms on a variety of control tasks. Our results show that MuJoCo is the best engine to transfer the learning to other engines. On the other hand, none of the algorithms generalize when trained on PyBullet. We also found out that various algorithms have a promising generalizability if the effect of random seeds can be minimized on their performance.
\end{abstract}


\section{Introduction}
\label{sec:introduction}

Simulation is one of the basic tools in robotics research \cite{torres2016} \cite{reckhaus2010overview}. Through simulation, it is possible to perform many runs of experiments without the constraints of a real robot system, such as parts failure, sequential runs, and safety \cite{dulac2019challenges}. Experiments in simulation can also be run faster than real time \cite{andrychowicz2020learning}, which helps in reducing the time constraints into the trial-and-error process of techniques like reinforcement learning.

But the quality of any task learned in simulation depends on the quality of the simulator \cite{collins2020traversing}, for example, in the numerical quality of the simulated dynamics or sensor modalities. Most agents learned in simulation do not generalize to reality, which is called simulation to reality gap \cite{mouret201720}. In some cases agents learned in simulation can be transferred to reality \cite{Hwangboeaau5872} \cite{kadian2019we}.

Reinforcement learning (RL) involves an agent taking actions and interacting with its environment to maximize long-term accumulated rewards. An RL agent learns how good or bad its actions are based on the rewards returned by the environment. In recent years, RL has been successful in a variety of domains including game playing and robotics, most of which involve continuous control tasks \cite{Hwangboeaau5872} \cite{Duan16}. Multiple iterations of training are needed before an RL algorithm can be used. A simulated environment that can correctly reflect the real world is needed to achieve the desired goal. An environment is nothing but the laws of physics that process the agents' actions and determine their consequences. Using reinforcement learning on a real robot without simulation is difficult and introduces additional learning problems \cite{mahmood2018setting} \cite{zhu2020ingredients}.

This work focuses on the tasks that use physics engines for simulating their environments. Several state of the art RL algorithms perform extremely well in physics-based environments \cite{Duan16}. These tasks are generally implemented in simulation using a particular physics engine. However, to the best of our knowledge, there is not much work done that deals with understanding the effect of physics engines on the performance of RL algorithms. This work aims to study the generalization of RL algorithms across multiple physics engines for continuous control tasks. This will help in understanding the algorithms better, evaluate their robustness \cite{chan2019}, and limitations in their learning capabilities. Algorithmic progress of tasks in simulation is not guaranteed to transfer into reality  \cite{kadian2019we}, so our work can be seen as a similar evaluation across different physics engines (simulated realities) using RL.

The work in this paper falls into the category of simulation-simulation gap \cite{collins2019benchmarking}, where it can be expected that agents transferred across simulators would not generalize, and this was our initial expectation, but our results show that in some cases transfer is possible across different physics engines, which is encouraging and calls for more research into understand the conditions needed for generalization in these cases.
Simulation in Robotics is of particular aid to produce varied samples for learning behaviors, so understanding the limitations of simulation and its transferability is also useful for robotics researchers and practitioners.

Our contributions are: we evaluate multiple reinforcement learning algorithms for continuous control on a set of common tasks while varying the physics engine implementing the environment. We find that in most cases there is a good degree of generalization across physics engines for some environments, and this does not depend on the reinforcement learning algorithm, but on the environment and task itself. We also evaluate the effect across multiple random seeds, and show that when the agent is able to transfer across physics engines, its performance also depends on the random seed to some extent, which is consistent with other results that show the variability of RL on random seeds like \cite{peter2017} and \cite{chan2019}.

We believe that these insights will motivate future studies of how generalization in reinforcement learning agents interact with the environment and its dynamics, and in bridging the simulation-reality gap.

\section{Related Work}
\label{sec:related}
To the best of our knowledge, there are no studies available that extensively compare the generalization of deep reinforcement learning algorithms across physics engines. In this section, we discuss prior research work that has been done in exploring the relationship between the physics engines and RL in general.

\cite{Duan16} provides a benchmark for continuous control tasks using a variety of RL algorithms including TRPO, Covariance Matrix Adaption Evolution Strategy (CMA-ES) \cite{hansen2001}, and DDPG. The authors highlight the importance of systematic evaluation of algorithms and encourage other researchers to evaluate their algorithms on the benchmark tasks.

The authors in \cite{collins2018} attempt to quantify the disparity of the performance of control systems between simulations and real-world tasks. This disparity is referred to as the \textit{reality gap}, which makes the solutions learned in simulation to not perform well in the real-world. \cite{collins2018} focuses on quantifying the reality gaps in robotic-grasping, which is relevant in a variety of domains like industrial robotics and assistive living. They compare different physics engines and simulators like Bullet, MuJoCo, ODE, Vortex, and Newton. The results suggest that combining multiple physics engines within a simulation for discrete periods can generate a model that reduces the reality gap. \cite{mouret201720} reviews different simulators for their simulation-reality gap. The authors suggest designing simulators that give a confidence score along with the prediction. \cite{Hwangboeaau5872} showed that the simulation-reality gap can be significantly reduced even for a complex robotic system by combining well-simulated systems with actuator models learned from reality. Similarly, \cite{andrychowicz2020learning} shows that policies learned entirely from simulation can still be successfully transferred to a physical robot.

On the other hand, \cite{torres2016} compares a variety of freely available simulators for mobile robots. The study compares Player-Stage-Gazebo (PSG), Open Dynamics Engine (ODE), Carnegie Mellon robot navigation toolkit (Carmen), and Microsoft Robotics Developer Studio (MRDS). Each simulator is compared for its features, ease of use, and the reliability in modeling the robot motion also taking the real-world physics into account. The simulators and the real robot were compared on their modeling ability of the robot motion on an uneven road. The results indicate that neither of the simulators is sufficient or superior to one another. Each simulator is suitable for particular domains. For example, Carmen is more suitable for navigation and motion planning. ODE should be used for tasks that require high physics accuracy. MRDS can be useful in modeling navigation and interaction strategies.
	
Similarly, \cite{chung2016} also compares different physics engines for predictable behavior in contact simulations. Their results show that ODE \cite{ode} and DART \cite{dart} produce patterns that are the most predictable. On the other hand, \cite{Erez2015} compares physics engines with respect to robotics. MuJoCo proved to be the most accurate and the fastest one. The tests also included a task (27 Capsules) more relevant to gaming engines, on which ODE proved to be the fastest. This showed that individual physics engines perform best on the tasks they are primarily designed for.

\section{Experimental Setup}
\label{sec:setup}
In this section, we describe the experimental setup used in this work. This includes the different tasks used to study the performance of the reinforcement learning algorithms. We also discuss the physics engines which were used to generate the tasks, network architecture, performance metrics and the methodology for the experiments.

\begin{figure}[t]
     \centering
     \begin{subfigure}[b]{0.15\textwidth}
         \centering
         \includegraphics[width=\textwidth]{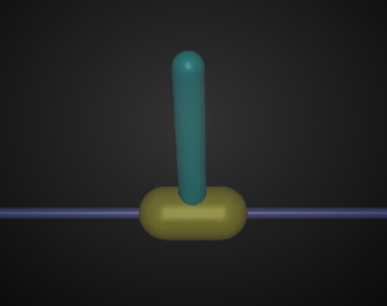}
         \caption{Inverted Pendulum}
         \label{fig:ip}
     \end{subfigure}
     \hfill
     \begin{subfigure}[b]{0.15\textwidth}
         \centering
         \includegraphics[width=\textwidth]{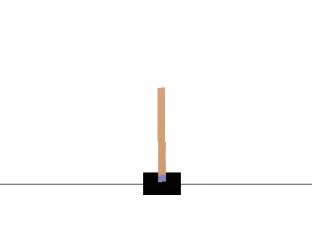}
         \caption{Cartpole}
         \label{fig:cartpole}
     \end{subfigure}
     \hfill
     \begin{subfigure}[b]{0.15\textwidth}
         \centering
         \includegraphics[width=\textwidth]{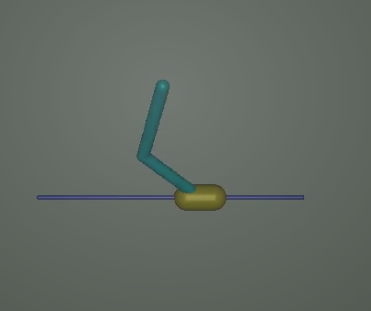}
         \caption{IDP}
         \label{fig:idp}
     \end{subfigure}
     \hfill
     \begin{subfigure}[b]{0.15\textwidth}
         \centering
         \includegraphics[width=\textwidth]{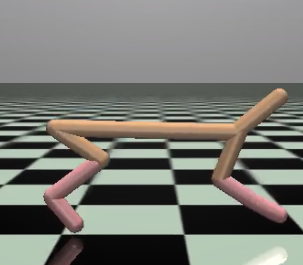}
         \caption{Half Cheetah}
         \label{fig:cheetah}
     \end{subfigure}
     \hfill
     \begin{subfigure}[b]{0.15\textwidth}
         \centering
         \includegraphics[width=\textwidth]{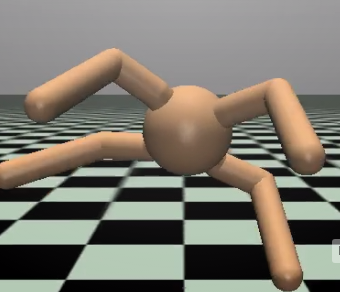}
         \caption{Ant}
         \label{fig:ant}
     \end{subfigure}
     \hfill
        \caption{Different continuous control tasks used for evaluation.}
        \label{fig:tasks}
	\end{figure}

\textbf{Tasks}. The tasks for evaluation were selected by taking the availability of their implementations across physics engines into consideration. Figure~\ref{fig:tasks} shows the different tasks and physics engine implementations used for evaluation .
	
\textbf{Physics Engines}. Physics engines are crucial in the development of reinforcement learning algorithms. They help in generating simulations that model the real-world problems. They reduce development time and also costs as the physical hardware is not required to test the algorithms. Three physics engines were used as part of this work along with the hard-coded dynamics from OpenAI Gym. MuJoCo \cite{Todorov2012} refers to Multi-Joint dynamics with Contact. It is a physics engine that facilitates fast and accurate simulation of systems with complex dynamics. Its applications span a variety of domains including robotics, gaming, animation, biomechanics to name a few. MuJoCo combines user convenience with computational efficiency with its layered design. The models are specified in an XML file format known as MJCF format which is easily readable and editable. It has some great features like simulation in generalized coordinates with modern contact dynamics, and unified treatment of different contacts like frictionless contacts, rolling and torsional friction contacts. Bullet \cite{bullet} is another physics engine that simulates real-time collision detection and models soft and rigid body dynamics. It is used for multiphysics simulation in robotics, machine learning, games, VR, and many other fields. PyBullet \cite{coumans2019} is an easy to use python module containing the python bindings for Bullet physics. Its main features include simulation of continuous and discrete collision detection, and support of two-way interaction between soft and rigid body dynamics. Similar to MuJoCo and Bullet, Open Dynamics Engine (ODE) \cite{ode} is a high performance, open-source library for modeling and simulating rigid body dynamics. ODE is used in many games, simulation tools and 3D authoring tools. It is also useful to simulate vehicles and other objects in VR environments.

\textbf{Methodology}. We compare the generalization performance of different RL algorithms across physics engines in the following way.

\begin{enumerate}
	\item Multiple implementations of a task based on different physics engines are identified.
	\item Each implementation of the task is trained using a deep RL algorithm. For example, all the three implementations (OpenAI Gym, ODE, Bullet) of the cartpole task  are trained using DQN and PPO.
	\item The trained models are evaluated on all the available implementations of the task including the one that it was trained on. This means, for cartpole, the model trained on OpenAI Gym is evaluated on ODE, Bullet and OpenAI Gym as well. The same applies to the models trained on ODE and Bullet.
	\item The steps 2 and 3 are repeated for a different RL algorithm on the same task.
	\item This process is repeated for different RL tasks.
\end{enumerate}

Table~\ref{table:task_summary} lists the combinations of tasks, RL algorithms and the physics engines used in this work.
	
\begin{table}[!t]
\centering
\begin{tabular}{@{}lll@{}}
\toprule
\textbf{Task}            & \textbf{Algorithms} & \textbf{Physics Engines} \\
\midrule
Cartpole                 & DQN, PPO & OpenAI Gym, ODE, Bullet  \\
Inverted Pendulum        & DDPG, TD3, SAC     & MuJoCo, Bullet           \\
Inverted Double Pendulum & DDPG, PPO           & MuJoCo, Bullet           \\
Half Cheetah             & VPG, TRPO          & MuJoCo, Bullet           \\
Ant                      & TD3, SAC            & MuJoCo, Bullet           \\
\bottomrule
\end{tabular}
\caption{Summary of the different combinations of tasks, algorithms and physics engines we used}
\label{table:task_summary}
\end{table}

\textbf{Random seeds}. Various studies suggest that deep reinforcement learning is brittle to random seeds \cite{peter2017}. This is also reinforced from our results. To avoid stochasticity, all experiments were run for 100 random seeds each. For each combination of a task and the algorithm, 100 models were generated, one for each random seed and the performance is averaged across different seeds. The random seeds selected were multiples of 10 i.e., 0, 10, 20, ..., 980, 990.

\textbf{Performance Metrics}. During training, performance refers to the average episode return for on-policy algorithms, and average test episode return for off-policy algorithms. The performance of each model is calculated for 100 epochs. As every combination of task and algorithm is run with 100 different random seed values, 100 different agents are trained. The training performance is then calculated by averaging performance of all the models over the course of training. During testing, each trained model is evaluated 100 times on all the implementations of the task. This is to reduce the stochasticity introduced by the random initial state. The testing performance is then calculated by averaging the performance for all the iterations per random seed value.

\textbf{Hyper-parameters}. All the algorithms used a network of size (64, 64) with relu units. The network structure was kept constant so that there is no bias introduced. The on-policy algorithms used 2000 steps of interaction per batch update. The off-policy algorithms used mini-batches of size 100 at each gradient descent step. Hyper-parameters were not tuned as the main objective of this project is to compare the generalization performance of the algorithms and not compare them against each other for the particular tasks. The discount rate used is $\gamma = 0.97$.

\textbf{Algorithmic Implementations}. Open source implementations of the algorithms have been used to ensure the consistency and robustness of our experiments. All the Implementations of algorithms except DQN used in this work have been taken from spinning up \cite{SpinningUp2018}, an educational resource by OpenAI. 
	

\section{Experimental Results and Analysis}
\label{sec:result}

The supplementary material contains all our experimental results in details. In this section we only show and analyze aggregate results (mean and standard deviation of test reward across RNG seeds).

\subsection{Cartpole}

\begin{table}[t]
\centering
\begin{tabular}{llll}
\toprule
\diagbox{\textbf{Train}}{\textbf{Test}}           & \textbf{Bullet}          & \textbf{Open AI Gym}      & \textbf{ODE}              \\
\midrule
\textbf{Bullet}       & 93.29 ${\displaystyle \pm}$ 44.16 & 10.39 ${\displaystyle \pm}$ 2.54  & 5.95 ${\displaystyle \pm}$ 2.28   \\
\textbf{Open AI Gym} & 8.92 ${\displaystyle \pm}$ 2.73   & 126.49 ${\displaystyle \pm}$ 67.61 & 66.9 ${\displaystyle \pm}$ 53.69   \\
\textbf{ODE}         & 11.14 ${\displaystyle \pm}$ 5.65  & 117.43 ${\displaystyle \pm}$ 78.07 & 112.28 ${\displaystyle \pm}$ 78.93 \\
\bottomrule
\end{tabular}

\caption{Comparison of DQN generalization on cartpole across physics engines.}
\label{table:cartpole dqn test}
\end{table}

As seen in Table~\ref{table:cartpole dqn test}, the high values of standard deviations show that random seeds play an important role in the performance of DQN. The generalizing ability of DQN is rather poor when trained on PyBullet based implementation and tested on the other two implementations. There is a better generalization performance seen when trained on the ODE based implementation and tested on that of OpenAI Gym. However, DQN generalizes the best from ODE to OpenAI Gym as the average performance is quite close, although the standard deviation is on the higher side. The models trained on PyBullet generate a maximum reward of less than 20 when tested on OpenAI Gym and ODE. On the other hand, the models trained on OpenAI Gym have a maximum reward of less than 20 on PyBullet but have a value of around 200 on ODE for around 10 of them. The models trained on ODE evaluate to a maximum reward of 50 on PyBullet but generate the maximum reward of around 200 for around 50 of them. This suggests that the generalization performance of DQN from PyBullet to ODE or OpenAI Gym is poor. This is also the case when transferring from OpenAI Gym or ODE to PyBullet. However, the generalization involving OpenAI Gym and ODE is highly dependent on the algorithm itself, in this case, DQN, as most of the models perform well on both the implementations.

\begin{table}[t]
\centering
\begin{tabular}{llll}
\toprule
\diagbox{\textbf{Train}}{\textbf{Test}}           & \textbf{Bullet}          & \textbf{Open AI Gym}      & \textbf{ODE}              \\
\midrule
\textbf{Bullet}       & 194.83 ${\displaystyle \pm}$ 12.79 & 25.5 ${\displaystyle \pm}$ 9.8 & 13.29 ${\displaystyle \pm}$ 4.49   \\
\textbf{Open AI Gym} & 18.1 ${\displaystyle \pm}$ 8.1  & 193.23 ${\displaystyle \pm}$ 13.64 & 75.63 ${\displaystyle \pm}$ 37.48   \\
\textbf{ODE}         & 22.53 ${\displaystyle \pm}$ 8.28  & 118.01 ${\displaystyle \pm}$ 66.15 & 195.0 ${\displaystyle \pm}$ 9.51 \\
\bottomrule
\end{tabular}
\caption{Comparison of PPO generalization on cartpole across physics engines}
\label{table:cartpole ppo test}
\end{table}

Table~\ref{table:cartpole ppo test} shows the generalization performance of PPO on cartpole task across PyBullet, OpenAI Gym, and ODE physics engines. The performance of the models trained on PyBullet is not very sensitive to random seeds as the performance variance across models is low. However the models do not generalize well on ODE and OpenAI Gym as the average reward earned is less than 50 in both the cases, whereas, on PyBullet itself, the maximum value of 200 is reached with the majority of the models. The models trained on OpenAI Gym perform poorly on PyBullet achieving an average reward of 18 and having the maximum value of less than 50. These models generalize better on ODE with an average value of more than 75 and reaching a maximum of around 200. Similar behavior is observed with the models trained on ODE. The generalization performance is poor on PyBullet with an average reward of 22 and a maximum of less than 50. However, the generalization is much better on OpenAI Gym with an average reward of 118 and a majority of them reaching 200. This reiterates the dependency of the generalization of RL agents on the random seeds.

\subsection{Inverted Pendulum}
\begin{table}[t]
\centering
\begin{tabular}{lll}
\midrule
\diagbox{\textbf{Train}}{\textbf{Test}} & \textbf{Bullet} & \textbf{MuJoCo} \\
\midrule
\textbf{Bullet} & 174.51 ${\displaystyle \pm}$ 48.67 & 5.97 ${\displaystyle \pm}$ 4.42 \\
\textbf{MuJoCo} & 104.93 ${\displaystyle \pm}$ 41.4   & 17.32 ${\displaystyle \pm}$ 25.11 \\
\bottomrule
\end{tabular}
\caption{Comparison of DDPG generalization on inverted pendulum across physics engines}
\label{table:ip ddpg test}
\end{table}

Table~\ref{table:ip ddpg test} shows the generalization performance of DDPG on the Inverted Pendulum task across PyBullet and MuJoCo. The models trained on PyBullet do not generalize well when evaluated on MuJoCo as the average reward generated is around 6 and the maximum reward is less than 25 across all the models. DDPG performs well when evaluated on PyBullet itself achieving an average reward of around 175 and a majority of the models reaching the maximum possible reward of 200 in this case. However, there is a high variance seen in the performance of the models across random seeds. The models trained on MuJoCo perform well when evaluated on PyBullet achieving an average reward of 105 and a few of the models reaching the value of 200. The performance is quite sensitive to random seeds as the reward varies from around 70 to 200. However, the performance of these models when evaluated on MuJoCo is quite poor with an average reward of less than 18 and the maximum reward value reached is around 125. The standard deviation is quite high which reiterates the effect of random seeds on the performance of DDPG in general.

\begin{table}[t]
\centering
\begin{tabular}{lllll}
\toprule
& \multicolumn{2}{c}{\textbf{TD3}} & \multicolumn{2}{c}{\textbf{SAC}}\\
\midrule
\diagbox{\textbf{Train}}{\textbf{Test}} & \textbf{Bullet} & \textbf{MuJoCo} & \textbf{Bullet} & \textbf{MuJoCo} \\ \midrule
\textbf{Bullet} & 182.03 ${\displaystyle \pm}$ 35.8 & 2.4 ${\displaystyle \pm}$ 0.9 & 152.75 ${\displaystyle \pm}$ 48.5 & 2.39 ${\displaystyle \pm}$ 0.36 \\
\textbf{MuJoCo} & 111.55 ${\displaystyle \pm}$ 33.06   & 198.94 ${\displaystyle \pm}$ 10.62 & 130.22 ${\displaystyle \pm}$ 30.6   & 199.49 ${\displaystyle \pm}$ 1.85 \\
\bottomrule
\end{tabular}
\caption{Comparison of TD3 and SAC generalization on inverted pendulum across physics engines}
\label{table:ip td3 & sac test}
\end{table}

Table~\ref{table:ip td3 & sac test} shows the generalization performance of SAC and TD3 on Inverted Pendulum task across PyBullet and MuJoCo. With TD3, the models trained on PyBullet do not generalize well on MuJoCo as the average reward earned across models is less than 3 and the maximum reward earned is less than 10. Their performance on PyBullet itself is quite sensitive to the random seeds. Although the average reward across models is greater than 180 and most of the models attain the maximum reward of 200, the values range from 50 to 200 which shows the high variance between models. On the other hand, the generalization performance of the models trained on MuJoCo is much better compared to that from PyBullet to MuJoCo. The models generate an average reward of around 112 and a couple of them even reaching 200. However, as seen in the plots, the performance is highly sensitive to random seeds as the rewards range from 75 to 200. The models perform extremely well when evaluated on MuJoCo itself. More than 95\% of the models generate a reward of 200 in this case which shows that their performance is not affected by random seeds on MuJoCo. Overall, the generalization performance of TD3 on Inverted Pendulum task is poor from PyBullet to MuJoCo but is much better from MuJoCo to PyBullet along with being sensitive to random seeds.

	Similar generalization behavior as TD3 is observed with SAC as well on the Inverted Pendulum task. The models trained on PyBullet poorly generalize on MuJoCo achieving an average reward value of around 2.5. None of the models generate a reward of more than 5. When these models are evaluated on PuBullet itself, an average reward of more than 150 is achieved. But the performance is highly affected by random seeds as the reward value ranges from 50 to 200. On the other hand, the generalization performance of SAC is much better from MuJoCo to PyBullet as the average reward of 130 and a maximum value of 175 is achieved. But the random seeds also play a huge role in the generalization performance as there is a huge variance across models. When the models trained on MuJoCo are tested on itself, there is a superior performance with an average value of almost 200 across models.

\subsection{Inverted Double Pendulum}

\begin{table}[t]
\centering
\begin{tabular}{lllll}
\toprule
    & \multicolumn{2}{c}{\textbf{DDPG}} & \multicolumn{2}{c}{\textbf{PPO}}\\
\midrule
\diagbox{\textbf{Train}}{\textbf{Test}} & \textbf{Bullet} & \textbf{MuJoCo} & \textbf{Bullet} & \textbf{MuJoCo} \\ \midrule
\textbf{Bullet} & 1487.78 ${\displaystyle \pm}$ 565.48 & 112.11 ${\displaystyle \pm}$ 35.75 & 1855.47 ${\displaystyle \pm}$ 19.86 & 94.99 ${\displaystyle \pm}$ 39.71\\
\textbf{MuJoCo} & 400.93 ${\displaystyle \pm}$ 164.29   & 1761.5 ${\displaystyle \pm}$ 385.51 & 1140.54 ${\displaystyle \pm}$ 276.31   & 1709.1 ${\displaystyle \pm}$ 71.27\\
\bottomrule
\end{tabular}
\caption{Comparison of DDPG and PPO generalization on inverted double pendulum across physics engines}
\label{table:idp test}
\end{table}

Table~\ref{table:idp test} shows the generalization performance of DDPG and PPO on Inverted Double Pendulum task. With DDPG, the models trained on PyBullet do not generalize well on MuJoCo as they achieve an average reward of 112 which is less than 10\% of what they achieve on PyBullet. The maximum reward value is also less than 200 across all the models on MuJoCo. The performance of the models is also highly sensitive to random seeds as the reward values range from 200 to 1800 even though a majority of them achieve closer to maximum possible reward. Similarly, the models generated on MuJoCo also perform poorly on PyBullet, although achieving a better average reward of around 400 when compared to that from PyBullet to MuJoCo. The performance of these models on MuJoCo is not very sensitive to random seeds as a majority of them achieve a high reward of around 1800. Overall, DDPG does not generalize well on Inverted Double Pendulum between PyBullet and MuJoCo.

Similar behavior is observed with PPO. The models trained on PyBullet generalize poorly on MuJoCo as they achieve an average reward of less than 100 which is around 5\% of the average reward on PyBullet. The random seeds also do not play a huge role as almost all the models have the reward value ranging from 10 to 250. The performance of these models is superior on PyBullet as almost all the models generate a reward value greater than 1750. The random seeds have very little effect on the performance. On the other hand, the models trained on MuJoCo promise a better performance as they achieve an average reward of greater than 1100 and a couple of them even reaching 1750. However, the effect of random seeds can be seen as the values range from around 650 to 1750. The performance of these models when evaluated on MuJoCo is much better as they achieve an average reward of greater than 1700 with a majority of them reaching more than 1500. The effect of random seeds is not significant as the standard deviation is quite low. Overall, the generalization performance of PPO on Inverted Double Pendulum task from PyBullet to MuJoCo is quite poor but is much better from MuJoCo to PyBullet, although being sensitive to random seeds.

\subsection{Half Cheetah}

\begin{table}[t]
\centering
\begin{tabular}{lllll}
\toprule
    & \multicolumn{2}{c}{\textbf{VPG}} & \multicolumn{2}{c}{\textbf{TRPO}}\\
\midrule
\diagbox{\textbf{Train}}{\textbf{Test}} & \textbf{Bullet} & \textbf{MuJoCo} & \textbf{Bullet} & \textbf{MuJoCo} \\ \midrule
\textbf{Bullet} & -485.95 ${\displaystyle \pm}$ 177.15 & -1793.86 ${\displaystyle \pm}$ 566.51 & -451.97 ${\displaystyle \pm}$ 414.73 & -1985.42 ${\displaystyle \pm}$ 756.62\\
\textbf{MuJoCo} & -878.71 ${\displaystyle \pm}$ 539.55   & -1003.59 ${\displaystyle \pm}$ 356.58 & 1153.18 ${\displaystyle \pm}$ 965.7   & -1490.46 ${\displaystyle \pm}$ 546.01 \\ 
\bottomrule
\end{tabular}
\caption{Comparison of VPG and TRPO generalization on half cheetah across physics engines}
\label{table:cheetah test}
\end{table}

Table~\ref{table:cheetah test} shows the generalization performance of VPG and TRPO on Half Cheetah task across PyBullet and MuJoCo. With VPG, the models generated on PyBullet do not generalize well when evaluated on MuJoCo as they achieve an average reward of around -1800 which is quite low compared to that on PyBullet which is around -500. The effect of random seeds is lower on PyBullet when compared to that on MuJoCo as the standard variation is quite high. On the other hand, the models trained on MuJoCo generalize well on PyBullet as they achieve comparable performance with that on MuJoCo itself. However, the variance in the reward is quite high across the models which indicates the sensitivity of VPG's performance to random seeds when trained on MuJoCo. Overall, it is seen that VPG generalizes and performs better on PyBullet when trained on MuJoCo.

The performance of TRPO is similar to that of VPG on Half Cheetah task. The models trained on PyBullet do not generalize well on MuJoCo as they achieve an average reward of around -1985 which is very low compared to that on PyBullet which is -450. The random seeds also play a significant role as there is a huge variance in reward values across the models. On the other hand, the models trained on MuJoCo generalize well on PyBullet as they achieve comparable performance, actually slightly better than that on MuJoCo itself. However, the performance is quite sensitive to random seeds as most of the reward values on PyBullet range from -500 to -3000 and some even reaching values greater than -4000. Similar behavior is seen on MuJoCo, although with a slightly lower variance across the models. Overall, TRPO generalizes well on Half Cheetah task from MuJoCo to PyBullet.

\subsection{Ant}

Table~\ref{table:ant test} shows the generalization performance of TD3 and SAC on Ant task across PyBullet and MuJoCo. With TD3, the models trained on PyBullet do not generalize on MuJoCo for Ant task as the average reward of -1400 on PyBullet is quite low when compared to that on MuJoCo which is around 2000. The performance is also affected by random seeds as seen by the high values of standard deviation. The poor generalization from PyBullet to MuJoCo on Ant is rather expected because the PyBullet version of the Ant consists of a majority of dummy variables. This limits the performance of the algorithm when all the variables are available in MuJoCo. On the other hand, the models trained on MuJoCo generalize on PyBullet to a great extent achieving around 60\% of the average reward on MuJoCo itself. The random seeds do not have a significant effect on the performance of these models as seen from the lower values of the standard deviation. Overall, TD3 generalizes well from MuJoCo to PyBullet for Ant.

The performance of SAC is similar to that of TD3. With SAC, the models generated on PyBullet do not generalize on MuJoCo as the average reward is quite low compared to that achieved on PyBullet itself. There is also very little effect of random seeds seen on the performance of SAC when trained on PyBullet. On the other hand, the models trained on MuJoCo achieve superior performance on PyBullet than on MuJoCo. They have an average reward of around 450 on PyBullet compared to around 140 on MuJoCo. The variance in the reward values across models is also low on PyBullet which suggests that SAC is not sensitive to random seeds when trained on MuJoCo. Overall, SAC has a promising generalization performance from MuJoCo to PyBullet.

\begin{table}[t]
    \centering
    \begin{tabular}{lllll}
        \toprule
        & \multicolumn{2}{c}{\textbf{TD3}} & \multicolumn{2}{c}{\textbf{SAC}}\\
        \midrule
        \diagbox{\textbf{Train}}{\textbf{Test}} & \textbf{Bullet} & \textbf{MuJoCo} & \textbf{Bullet} & \textbf{MuJoCo} \\ \midrule
        \textbf{Bullet} & 2012.33 ${\displaystyle \pm}$ 444.44 & -1406.56 ${\displaystyle \pm}$ 933.56 & 956.15 ${\displaystyle \pm}$ 124.88 & -154.65 ${\displaystyle \pm}$127.36\\ 
        \textbf{MuJoCo} & 345.04 ${\displaystyle \pm}$ 86.65   & 599.89 ${\displaystyle \pm}$ 266.25 & 448.03 ${\displaystyle \pm}$ 74.92   & 137.52 ${\displaystyle \pm}$ 107.37\\
        \bottomrule
    \end{tabular}
    
    \caption{Comparison of TD3 and SAC generalization on ant across physics engines}
    \label{table:ant test}
    \vspace*{-2em}
\end{table}

\subsection{Discussion and Analysis}

Overall we see that in many cases an agent learned using reinforcement learning does not generalize if trained in one physics engine, and tested in a different one, even as they in theory implement the same dynamics. This could be due to differences on how the environment dynamics are implemented inside each physics engine, as its the most likely source of error. This is particularly noticeable on cartpole trained on bullet and tested on gym and ode, and inverted pendulum trained on bullet and tested on mujoco.

But what was unexpected is that generalization is possible for some tasks and environments across physics engines, namely for cartpole on gym to/from ode, inverted pendulum on mujoco to bullet, inverted double pendulum on mujoco to bullet, half cheetah on mujoco to/from bullet, and ant on mujoco to bullet, with some small loss in performance but not catastrophically failing. In general this generalization does not seem to depend on the reinforcement learning algorithm that is being used (minus differences in performance across algorithms), only on the task and environment implementation.

When generalization across physics engines is possible, it does not seem to be stable, with performance heavily depending on the random seed used to train or test the agent/environment, indicating the need to disentangle the performance of a reinforcement learning agent from the random seed is a promising direction \cite{peter2017}.



\section{Conclusions and Future Work}
\label{sec:conclusion}

In this paper we have evaluated different reinforcement learning algorithms for continuous control, across different tasks and physics engine implementations of those tasks. We find that in most cases an agent learned using a certain physics engine fails to perform in the same task implemented with a different physics engine. This is expected, but this does not always happen. In some cases, across different random seeds, performance can reach or even exceed the level of the source physics engine, which encourages that the agent is able to generalize.


Generalization across physics engines does not seem to work like a binary (on/off) decision, with a varied spectrum in between able to generalize (on) and failure to generalize (off), specially when different random seeds are taken in to account.


We expect that our results will motivate new research into understanding what characterizes the differences between physics engines and how it related to the performance of reinforcement learning agents, specially used for robotics applications.



\clearpage


\bibliographystyle{plain}
\bibliography{bibliography.bib}  

\clearpage
\FloatBarrier
\section{Additional Training and Comparison Plots for Each Environment}

In this section we include additional results that did not fit in the main paper. These include one sample training curve for each reinforcement learning agent, and a visual comparison of all training runs and evaluations across physics engines, all separated by task/environment.

When evaluating generalization across physics engines, we make a matrix of plots, where in each row an agent is trained on a specific environment with a particular physics engine, and across columns the trained agent is evaluated in the same environment but implemented using a different physics engine. This is repeated with different physics engines used to train the agent and for different tasks/environments.

Each plot inside a matrix presents the average reward obtained for an episode, where the X axis represents individual trials (100 in total per plot), each with a different random seed, and the Y axis presents the average reward return obtained in that episode with a particular physics engine.

We also include at the beginning of each environment section a comparison of training runs across episodes, tested on the training physics engine, as a way to compare differences in learning performance across different physics engines and reinforcement learning algorithms.

\subsection{Cartpole}

\begin{figure}[!ht]
     \centering
     \begin{subfigure}[b]{0.6\textwidth}
         \centering
         \includegraphics[width=\textwidth]{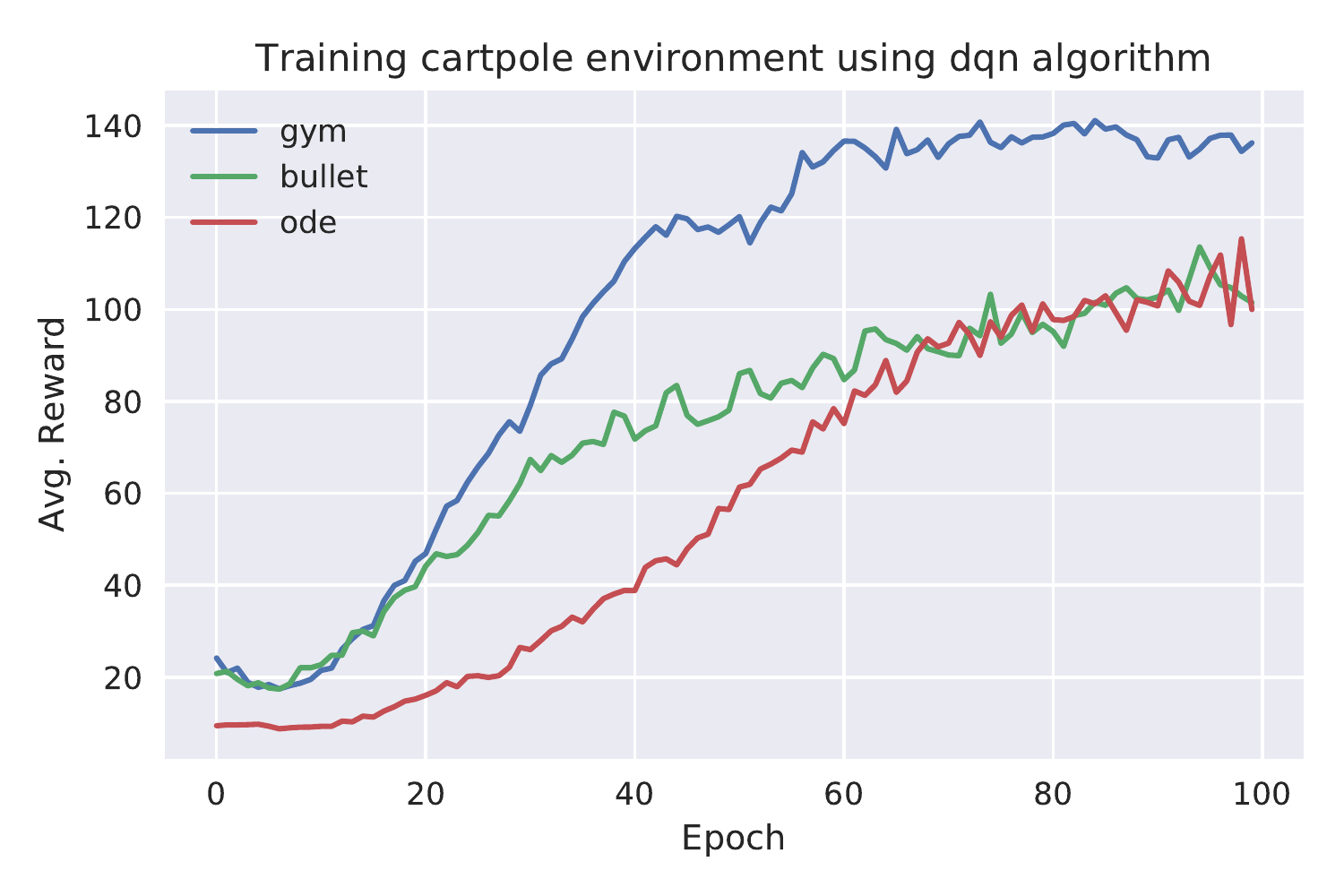}
         \caption{}
         \label{fig:cartpole dqn train}
     \end{subfigure}
     \hfill
     \begin{subfigure}[b]{0.6\textwidth}
         \centering
         \includegraphics[width=\textwidth]{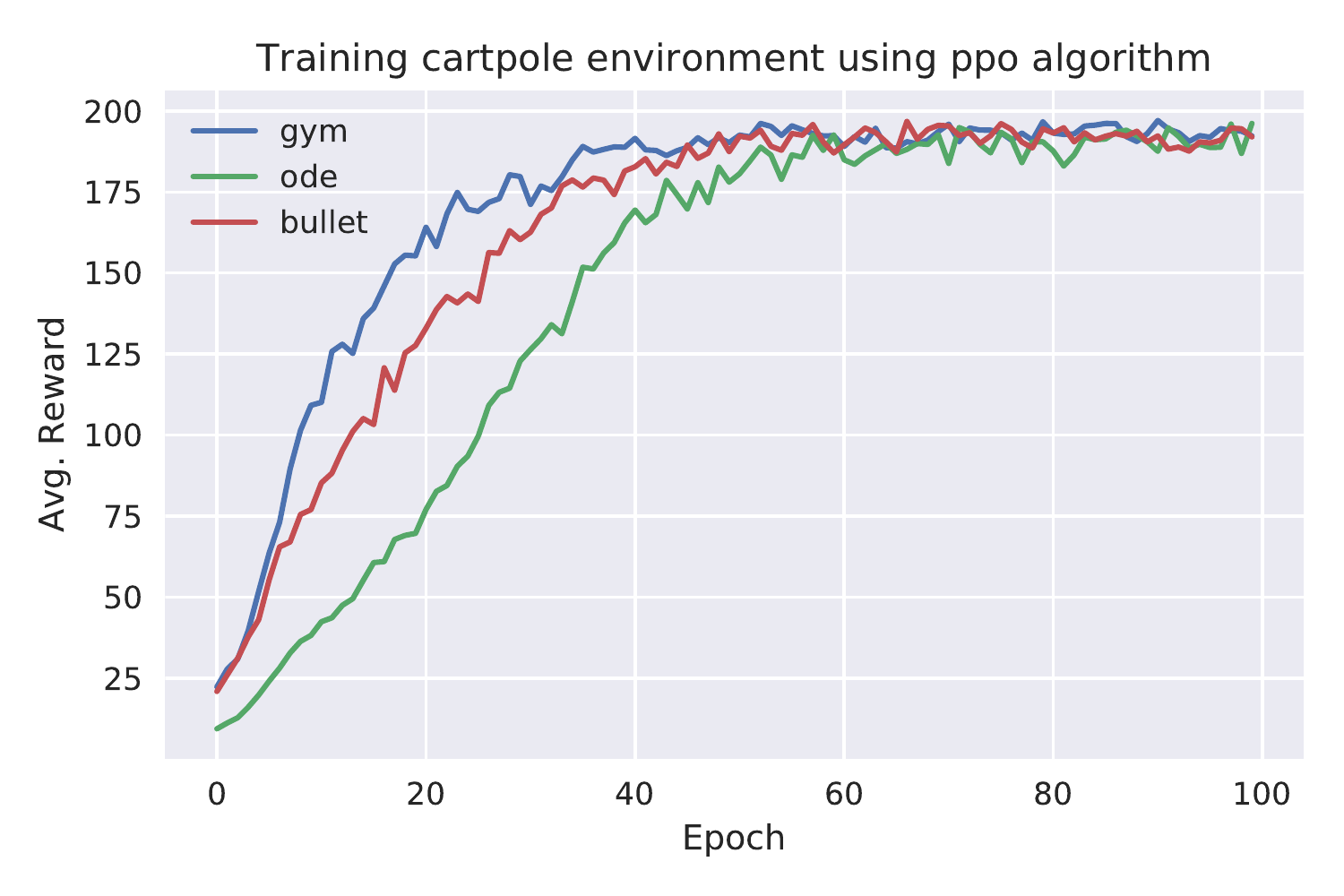}
         \caption{}
         \label{fig:cartpole ppo train}
     \end{subfigure}
        \caption{Performance as a function of average reward over 100 different random seeds during the course of training. (a) Performance of DQN on cartpole task across different physics engines. (b) Performance of PPO on cartpole task across different physics engines.}
        \label{fig:cartpole training}
\end{figure}

\begin{figure}[H]
		\centering
		\includegraphics[width=0.75\textwidth]{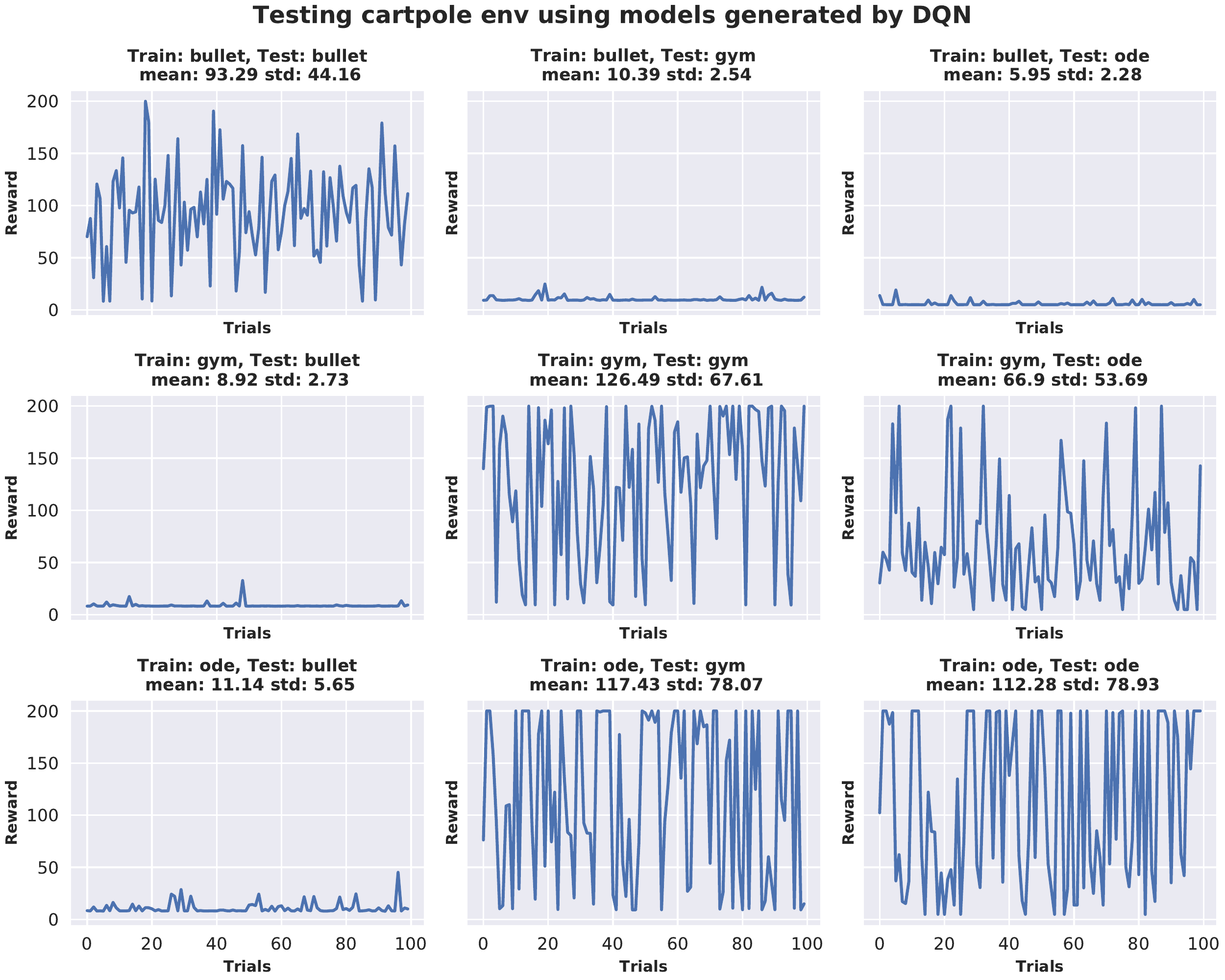}
		\caption{Comparison of generalization of DQN on cartpole task across physics engines}
		\label{fig:cartpole dqn test}
	\end{figure}

	\begin{figure}[H]
		\centering
		\includegraphics[width=0.75\textwidth]{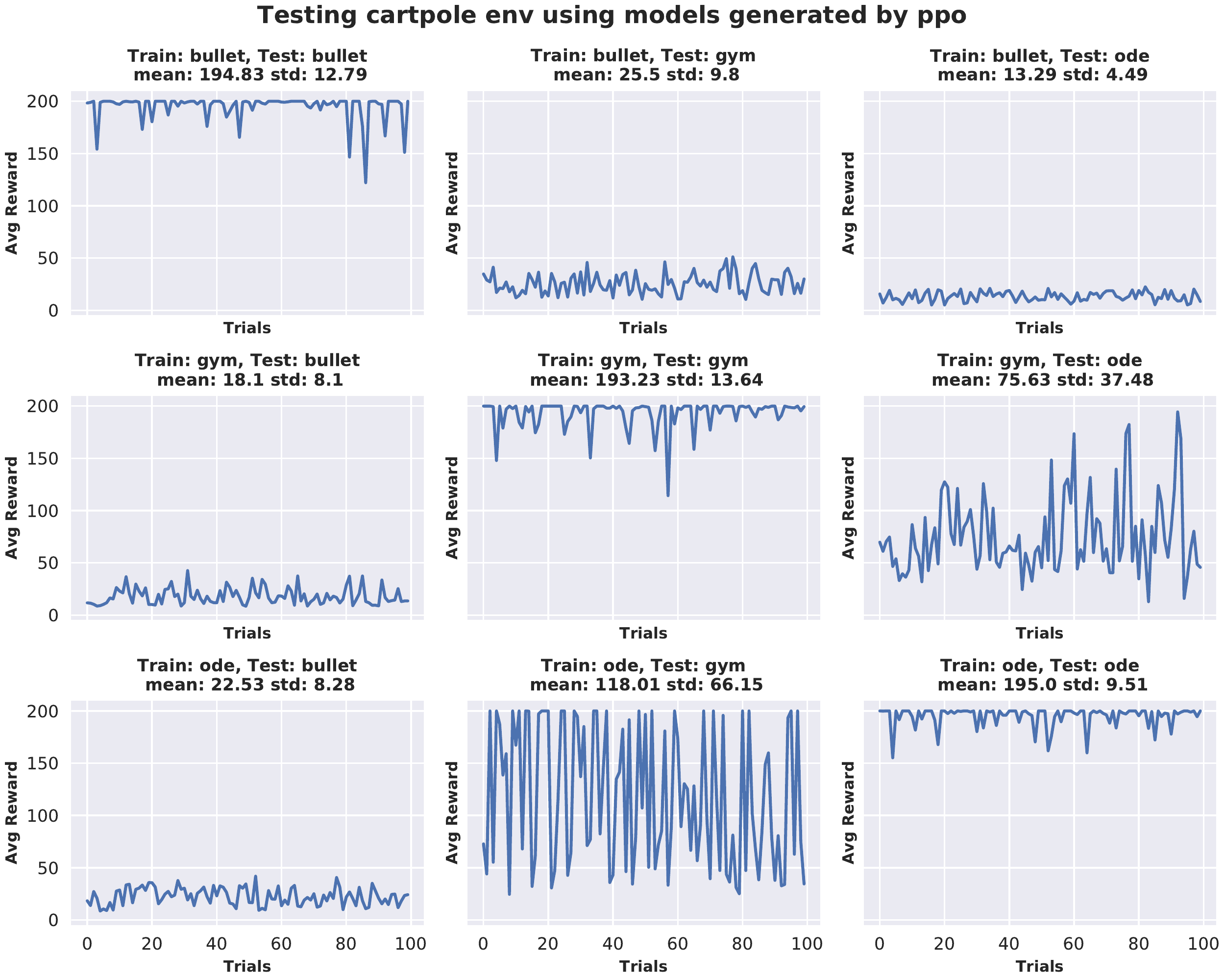}
		\caption{Comparison of generalization of PPO on cartpole task across physics engines}
		\label{fig:cartpole ppo test}
	\end{figure}

\subsection{Inverted Pendulum}
	
	\begin{figure}[H]
     \centering
     \begin{subfigure}[b]{0.6\textwidth}
         \centering
         \includegraphics[width=\textwidth]{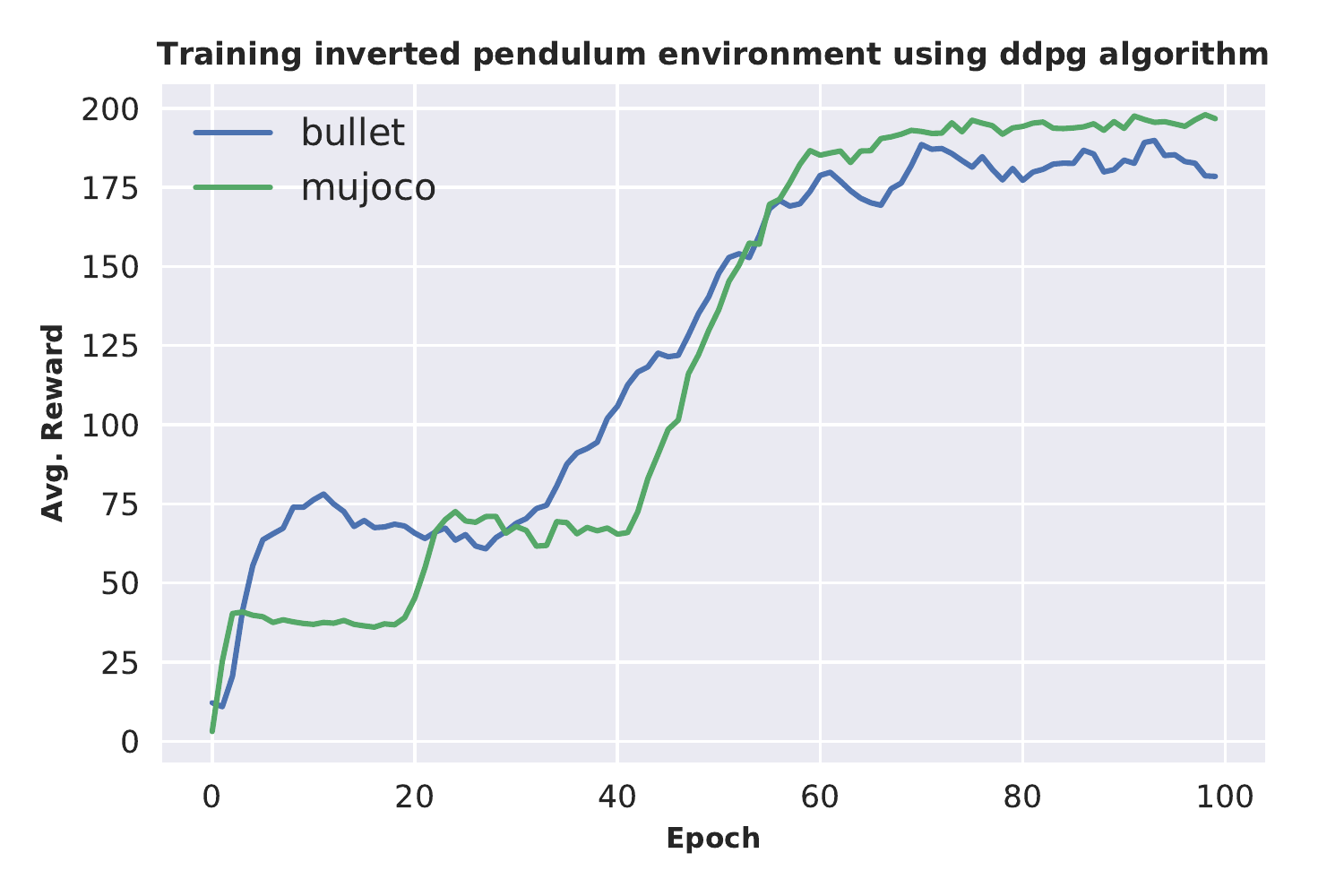}
         \caption{}
         \label{fig:ip ddpg train}
     \end{subfigure}
     \hfill
     \begin{subfigure}[b]{0.6\textwidth}
         \centering
         \includegraphics[width=\textwidth]{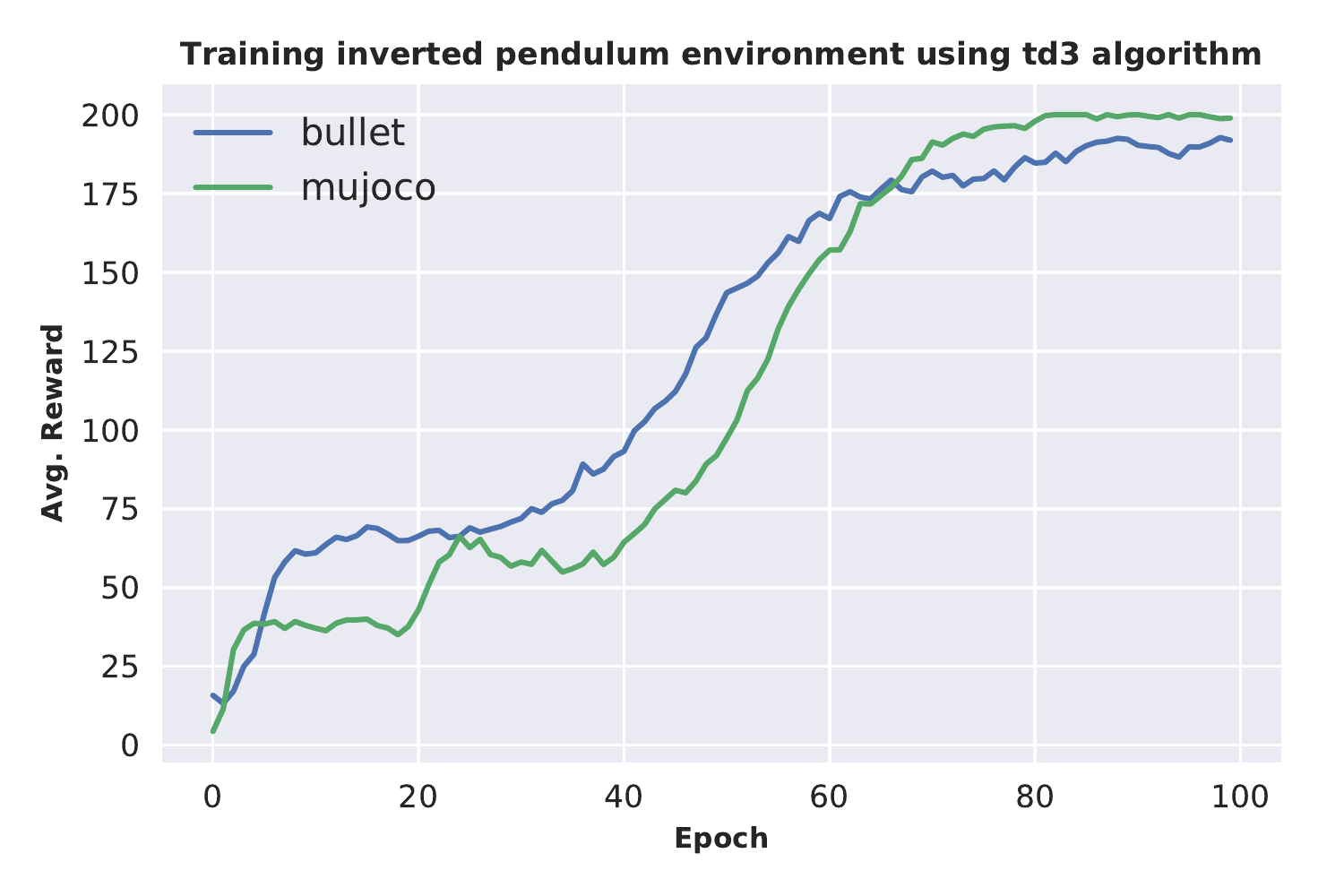}
         \caption{}
         \label{fig:ip td3 train}
     \end{subfigure}
     \hfill
     \begin{subfigure}[b]{0.6\textwidth}
         \centering
         \includegraphics[width=\textwidth]{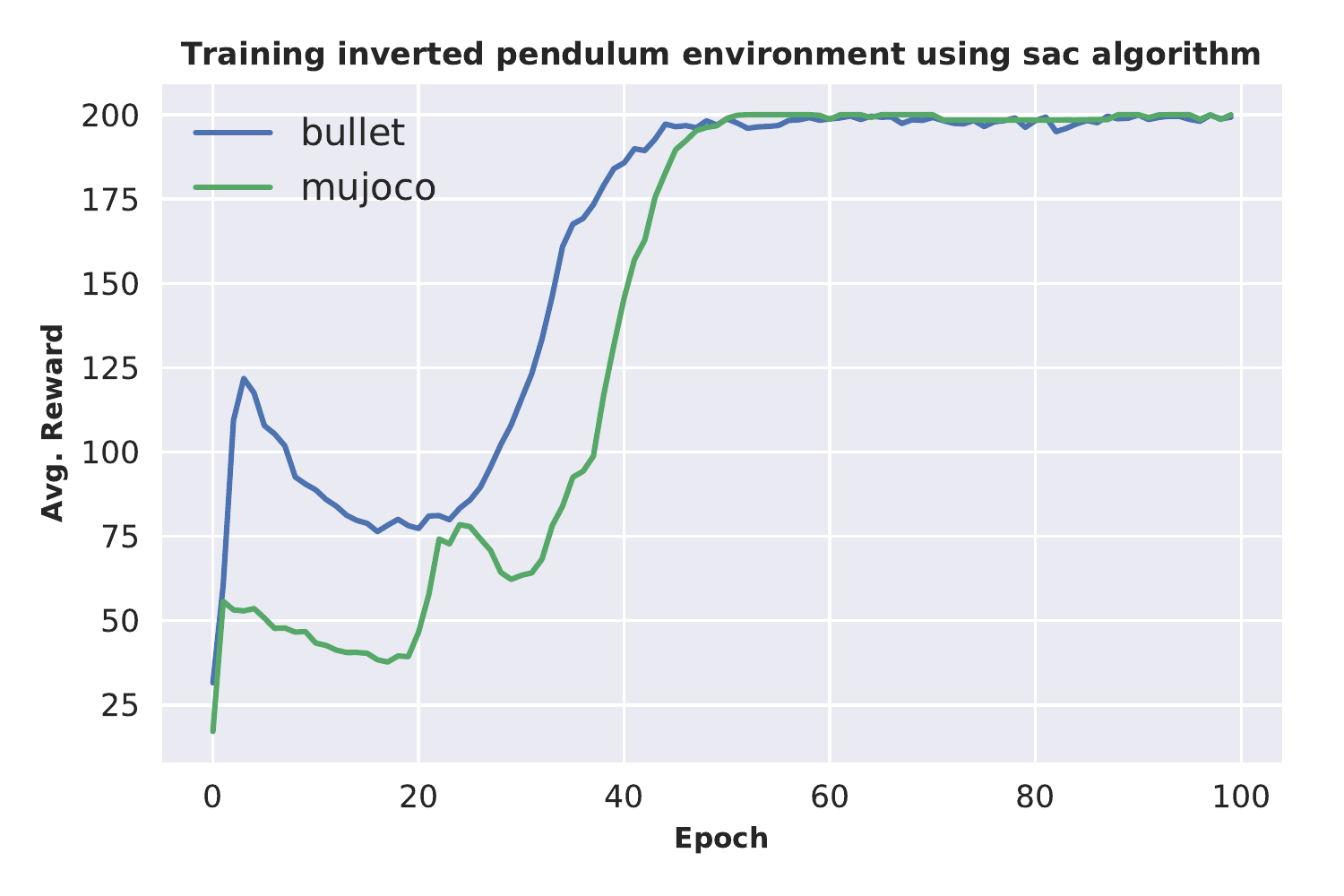}
         \caption{}
         \label{fig:ip sac train}
     \end{subfigure}
        \caption{Performance as a function of average reward over 100 different random seeds during the course of training. (a) Performance of DDPG on inverted pendulum task across different physics engines. (b) Performance of TD3 on inverted pendulum task across different physics engines. (c) Performance of SAC on inverted pendulum task across different physics engines.}
        \label{fig:ip training}
	\end{figure}
	
	\begin{figure}[H]
		\centering
		\includegraphics[width=0.75\textwidth]{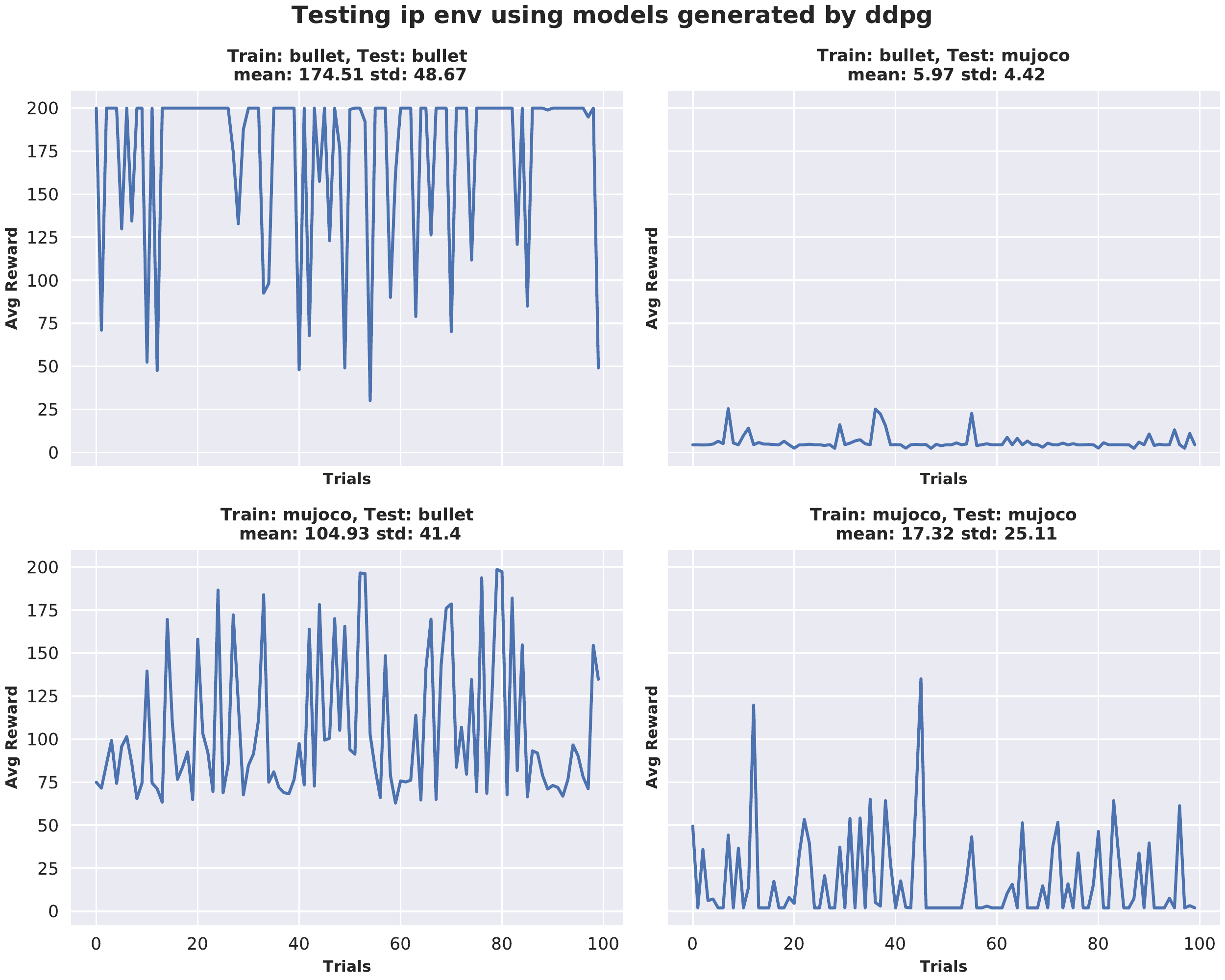}
		\caption{Comparison of generalization of DDPG on inverted pendulum task across physics engines}
		\label{fig:ip ddpg test}
	\end{figure}
	
	\begin{figure}[H]
		\centering
		\includegraphics[width=0.75\textwidth]{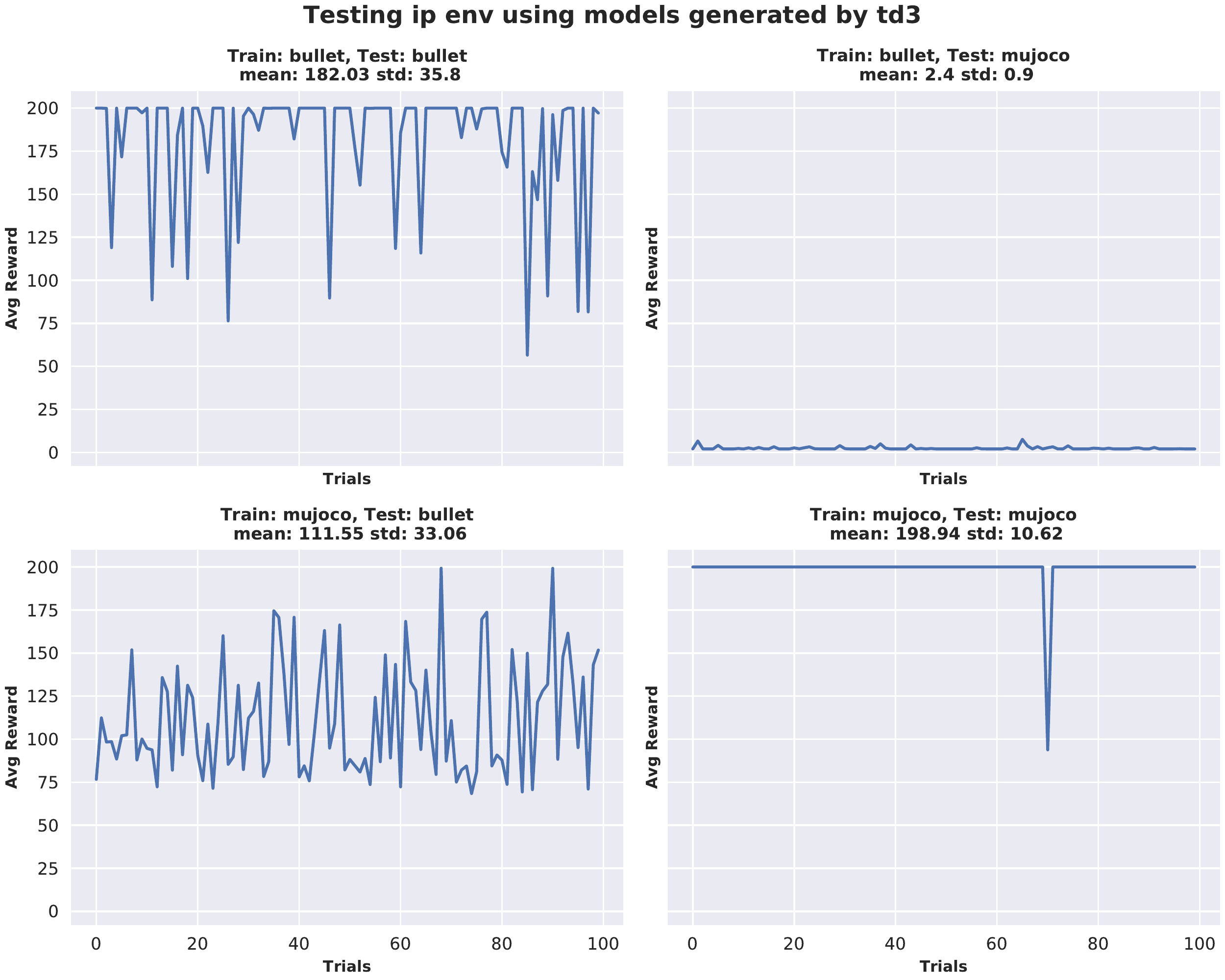}
		\caption{Comparison of generalization of TD3 on inverted pendulum task across physics engines}
		\label{fig:ip td3 test}
	\end{figure}
	
	\begin{figure}[H]
		\centering
		\includegraphics[width=0.75\textwidth]{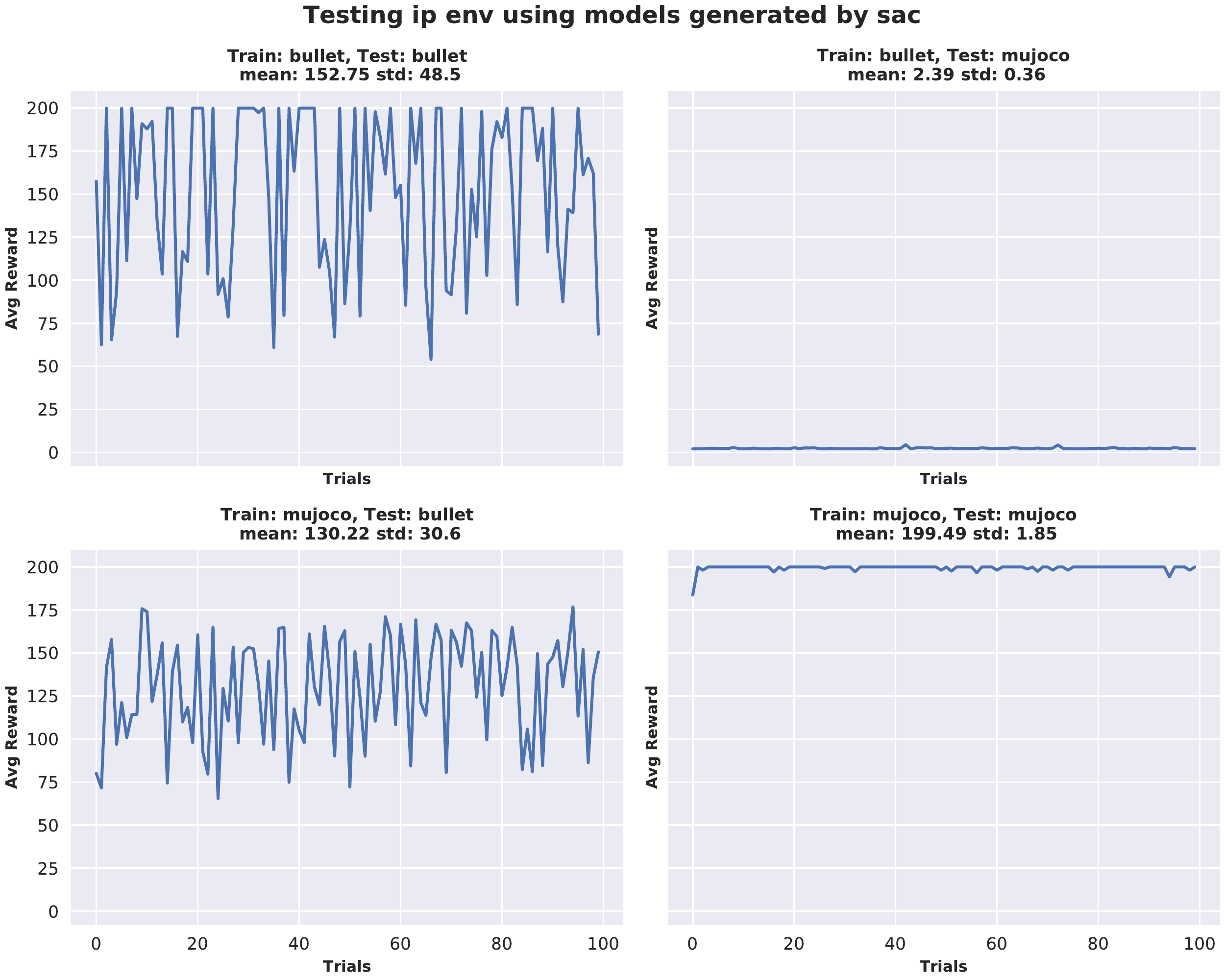}
		\caption{Comparison of generalization of SAC on inverted pendulum task across physics engines}
		\label{fig:ip sac test}
	\end{figure}

\clearpage
\subsection{Inverted Double Pendulum}
	
	\begin{figure}[!h]
     \centering
     \begin{subfigure}[b]{0.6\textwidth}
         \centering
         \includegraphics[width=\textwidth]{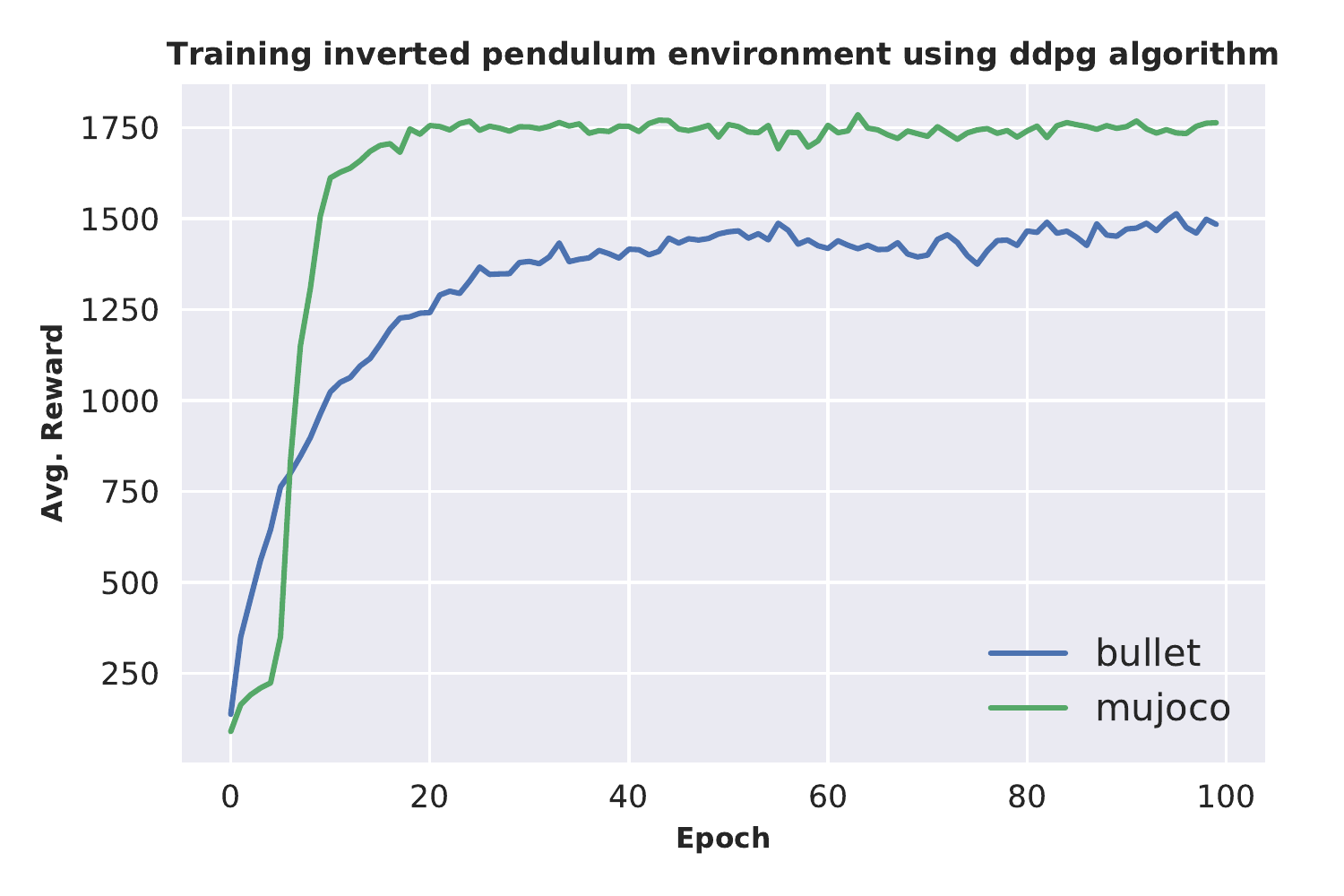}
         \caption{}
         \label{fig:idp ddpg train}
     \end{subfigure}
     \hfill
     \begin{subfigure}[b]{0.6\textwidth}
         \centering
         \includegraphics[width=\textwidth]{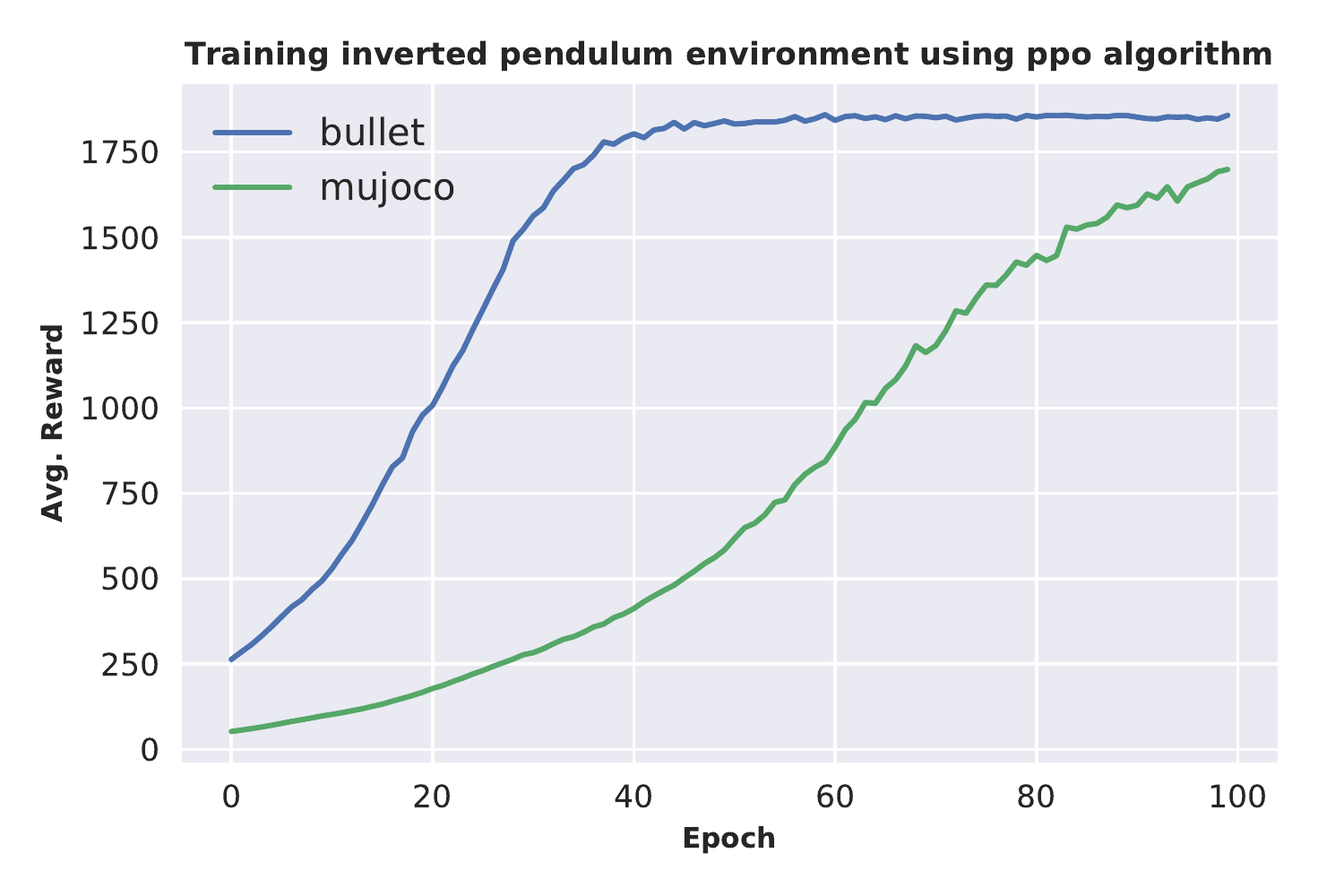}
         \caption{}
         \label{fig:idp ppo train}
     \end{subfigure}
        \caption{Performance as a function of average reward over 100 different random seeds during the course of training. (a) Performance of DDPG on inverted double pendulum task across different physics engines. (b) Performance of PPO on inverted double pendulum task across different physics engines.}
        \label{fig:idp training}
	\end{figure}
	
	\begin{figure}[!h]
		\centering
		\includegraphics[width=0.75\textwidth]{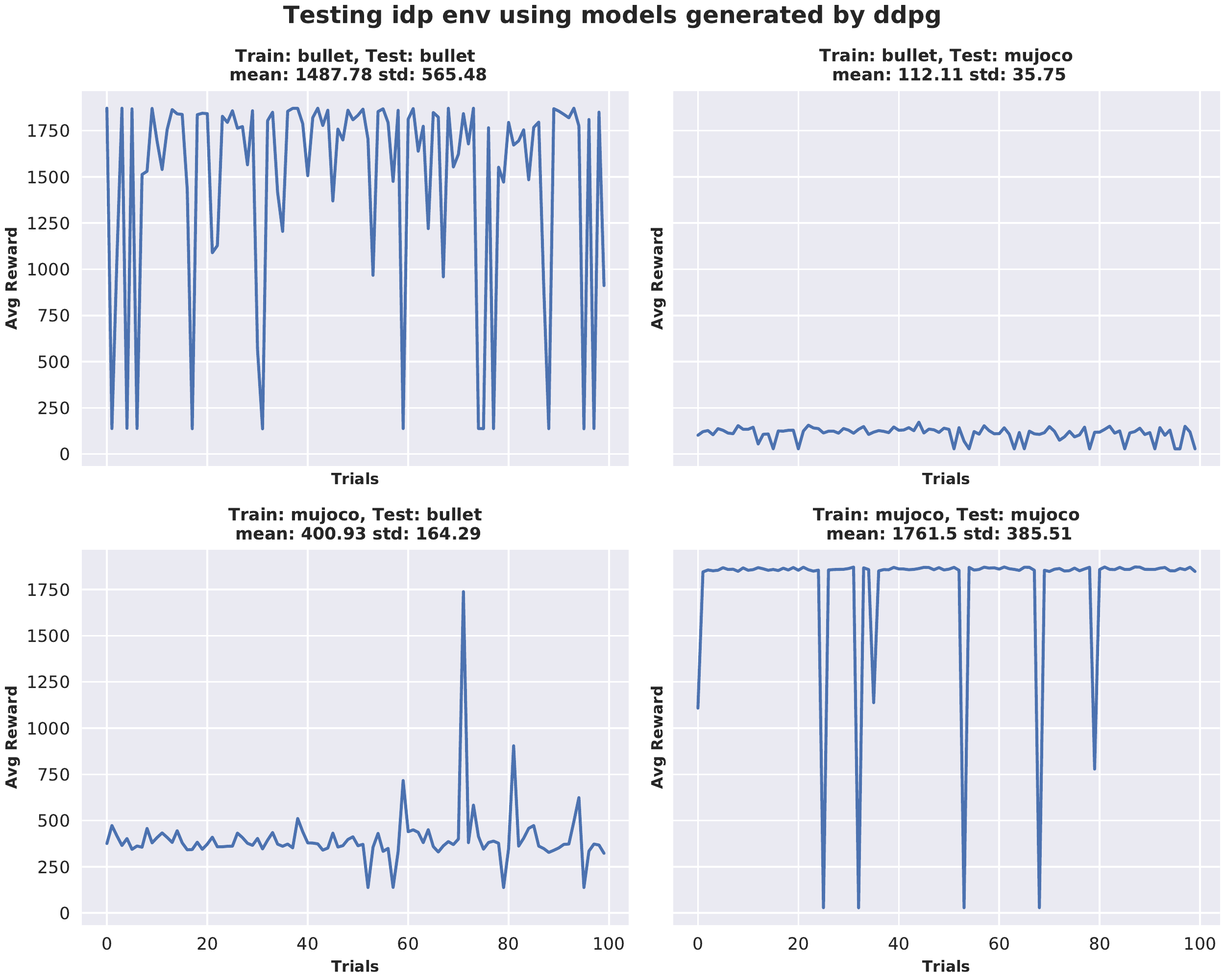}
		\caption{Comparison of generalization of DDPG on inverted double pendulum task across physics engines}
		\label{fig:idp ddpg test}
	\end{figure}
	
	\begin{figure}[!h]
		\centering
		\includegraphics[width=0.75\textwidth]{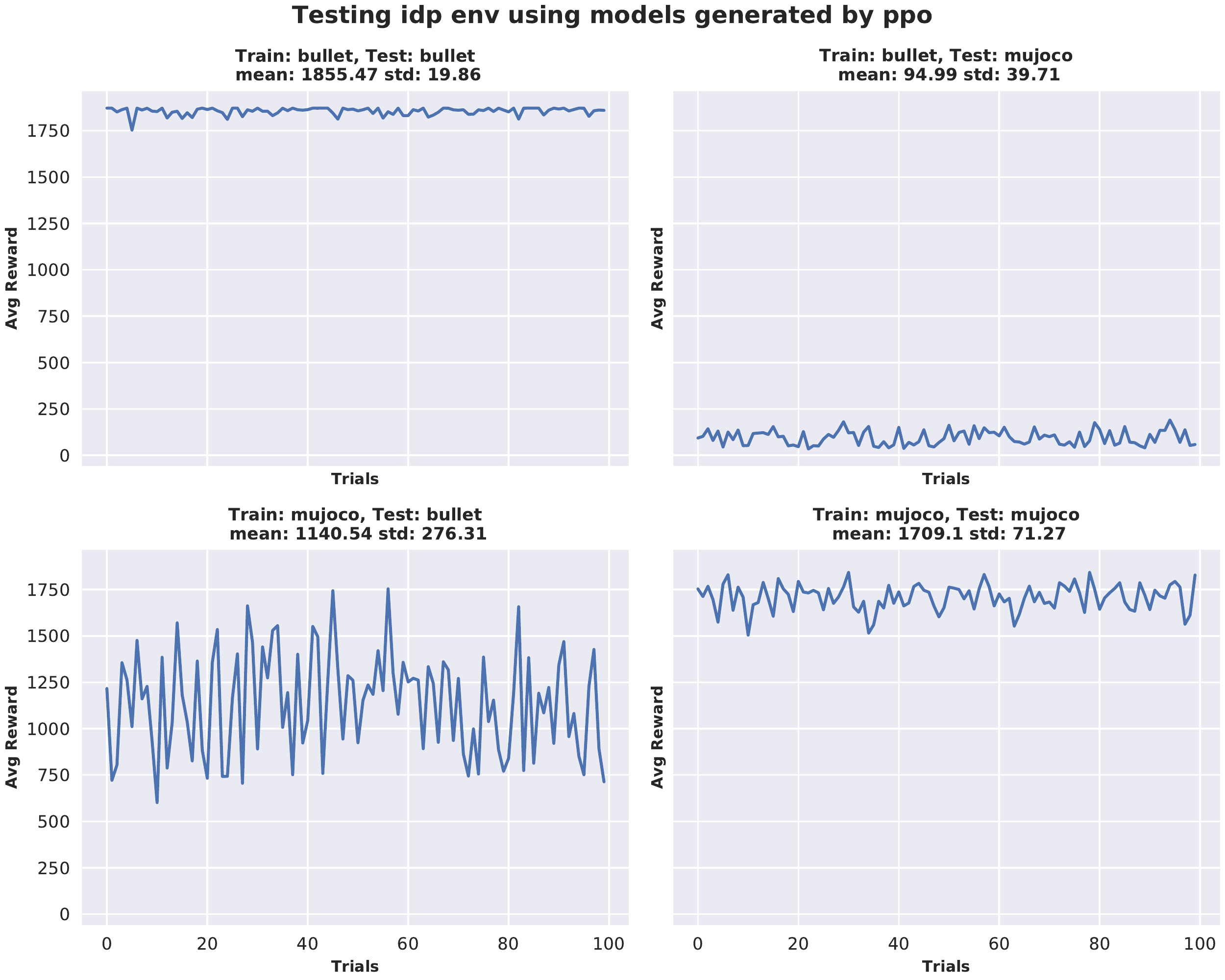}
		\caption{Comparison of generalization of PPO on Inverted Double Pendulum task across physics engines}
		\label{fig:idp ppo test}
	\end{figure}

\clearpage
\subsection{Half Cheetah}
	
	\begin{figure}[!h]
     \centering
     \begin{subfigure}[b]{0.6\textwidth}
         \centering
         \includegraphics[width=\textwidth]{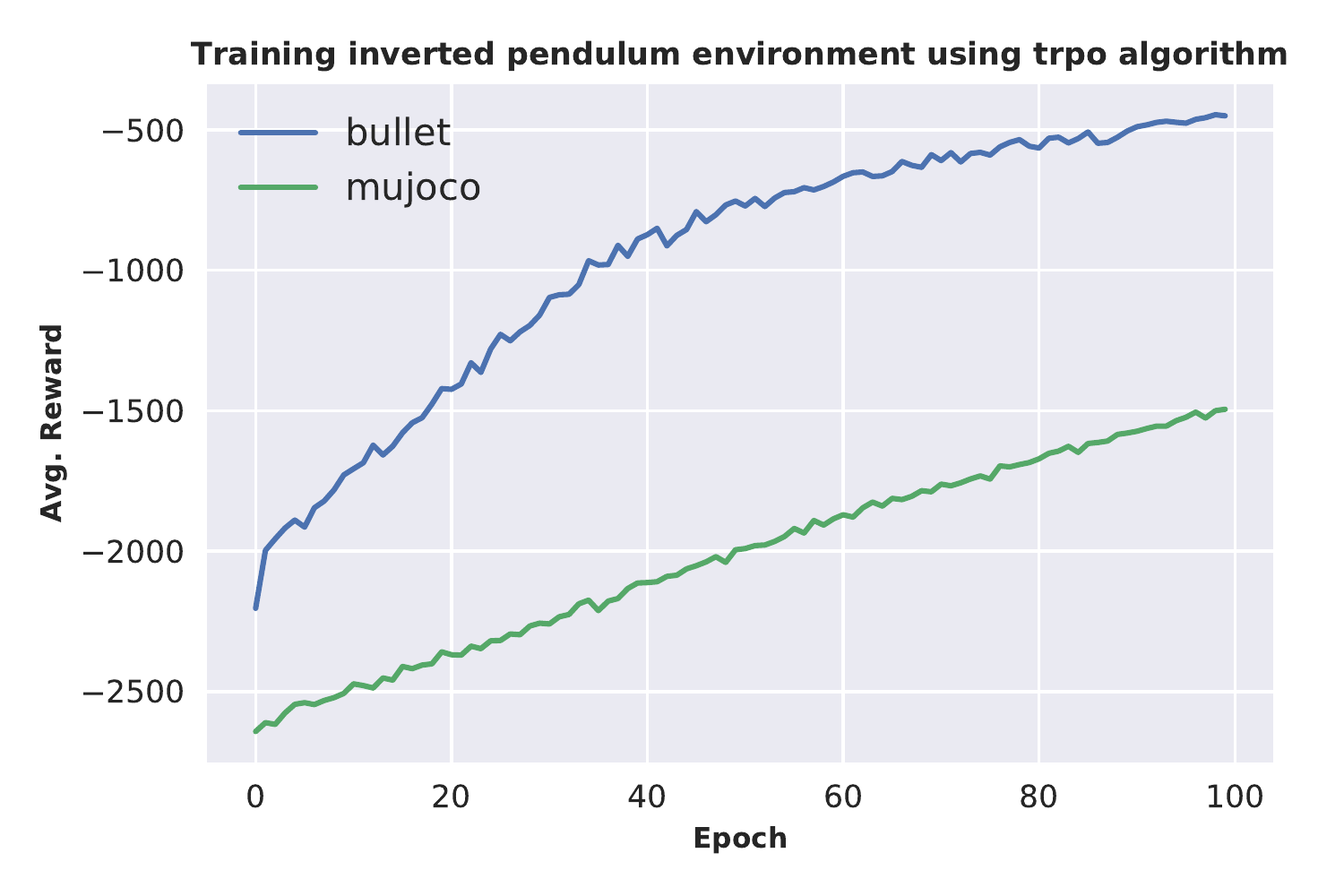}
         \caption{}
         \label{fig:cheetah trpo train}
     \end{subfigure}
     \hfill
     \begin{subfigure}[b]{0.6\textwidth}
         \centering
         \includegraphics[width=\textwidth]{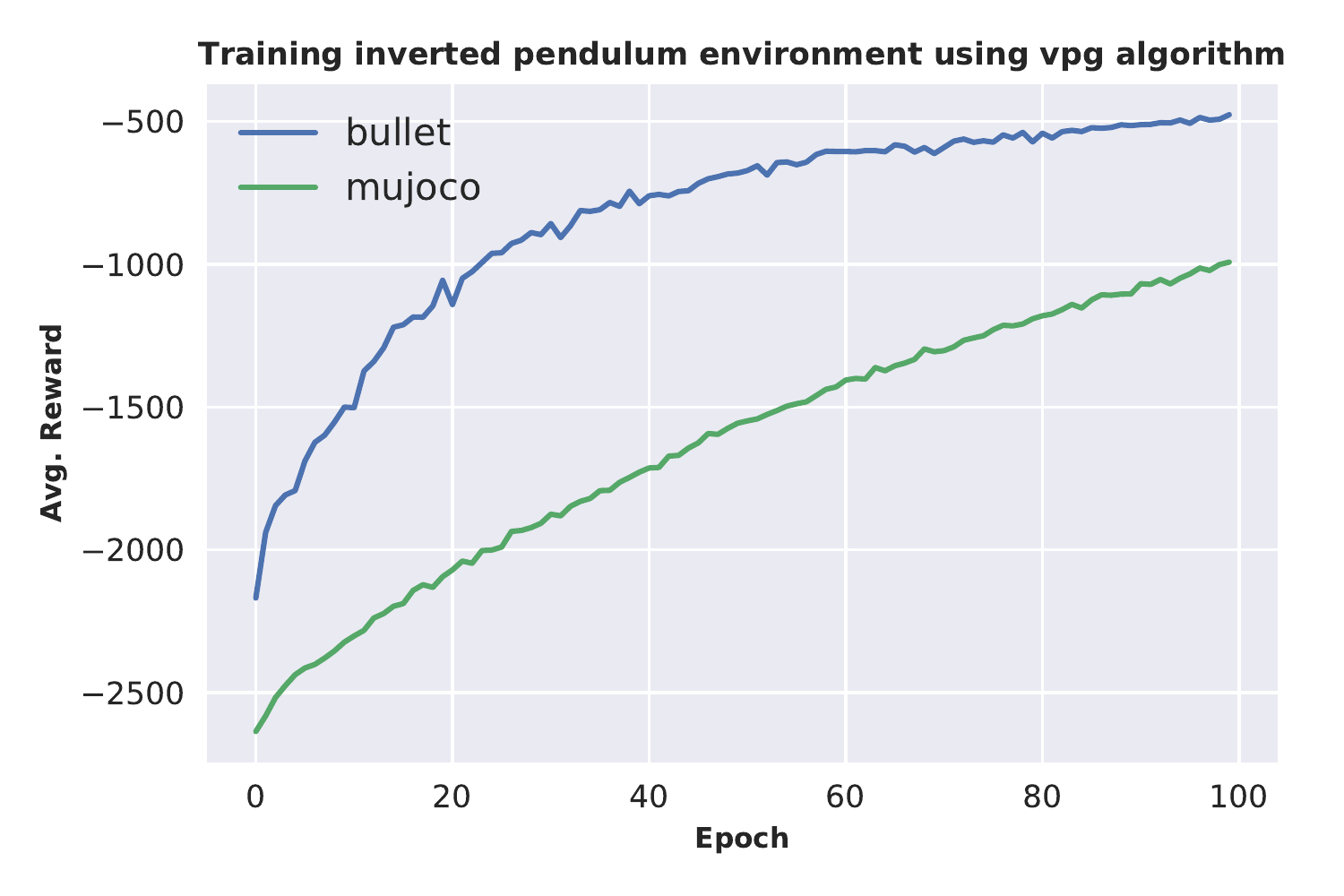}
         \caption{}
         \label{fig:cheetah vpg train}
     \end{subfigure}
        \caption{Performance as a function of average reward over 100 different random seeds during the course of training. (a) Performance of TRPO on half cheetah task across different physics engines. (b) Performance of VPG on half cheetah task across different physics engines.}
        \label{fig:cheetah training}
	\end{figure}
	
	\begin{figure}[!h]
		\centering
		\includegraphics[width=0.75\textwidth]{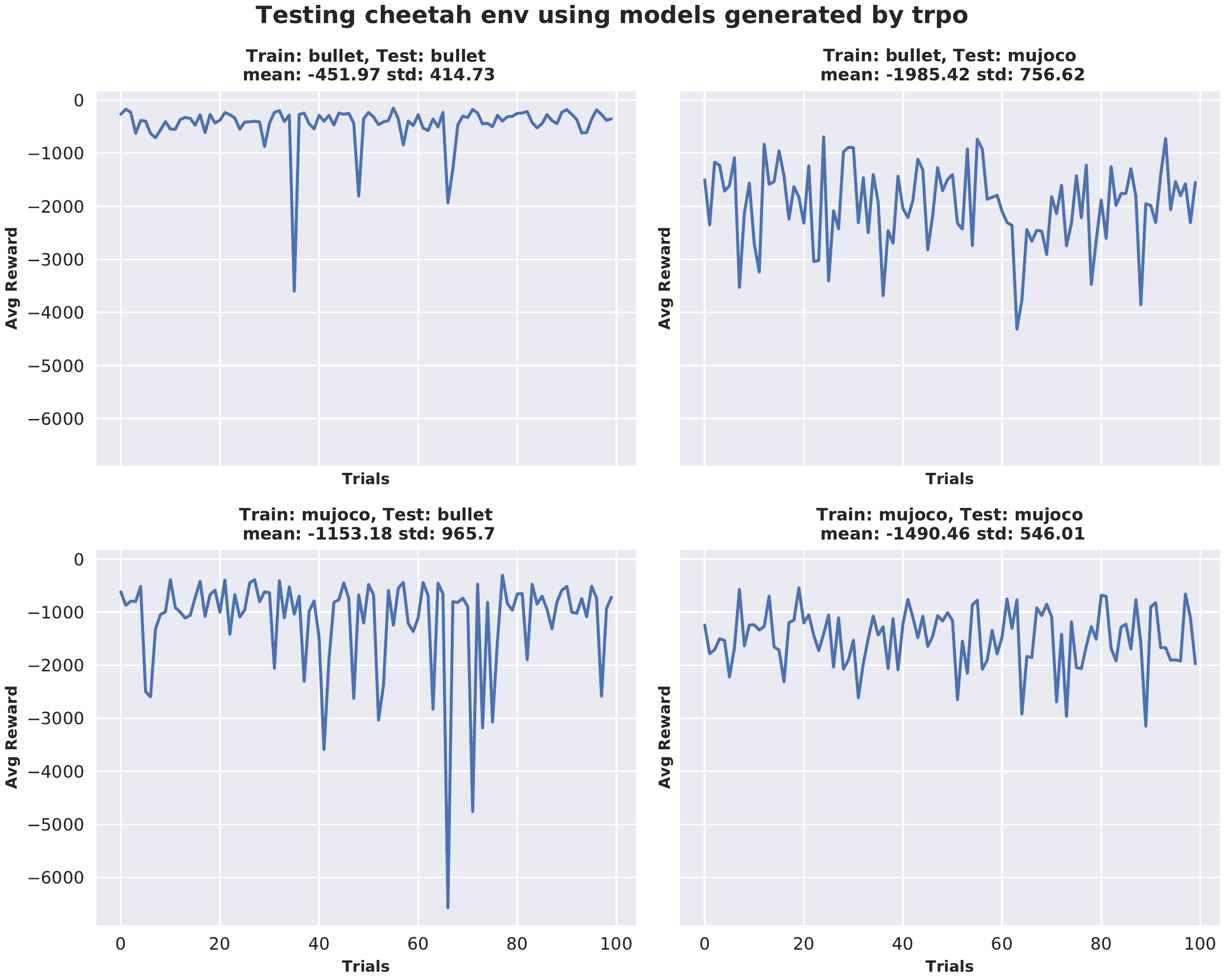}
		\caption{Comparison of generalization of TRPO on half cheetah task across physics engines}
		\label{fig:cheetah trpo test}
	\end{figure}
	
	\begin{figure}[!h]
		\centering
		\includegraphics[width=0.75\textwidth]{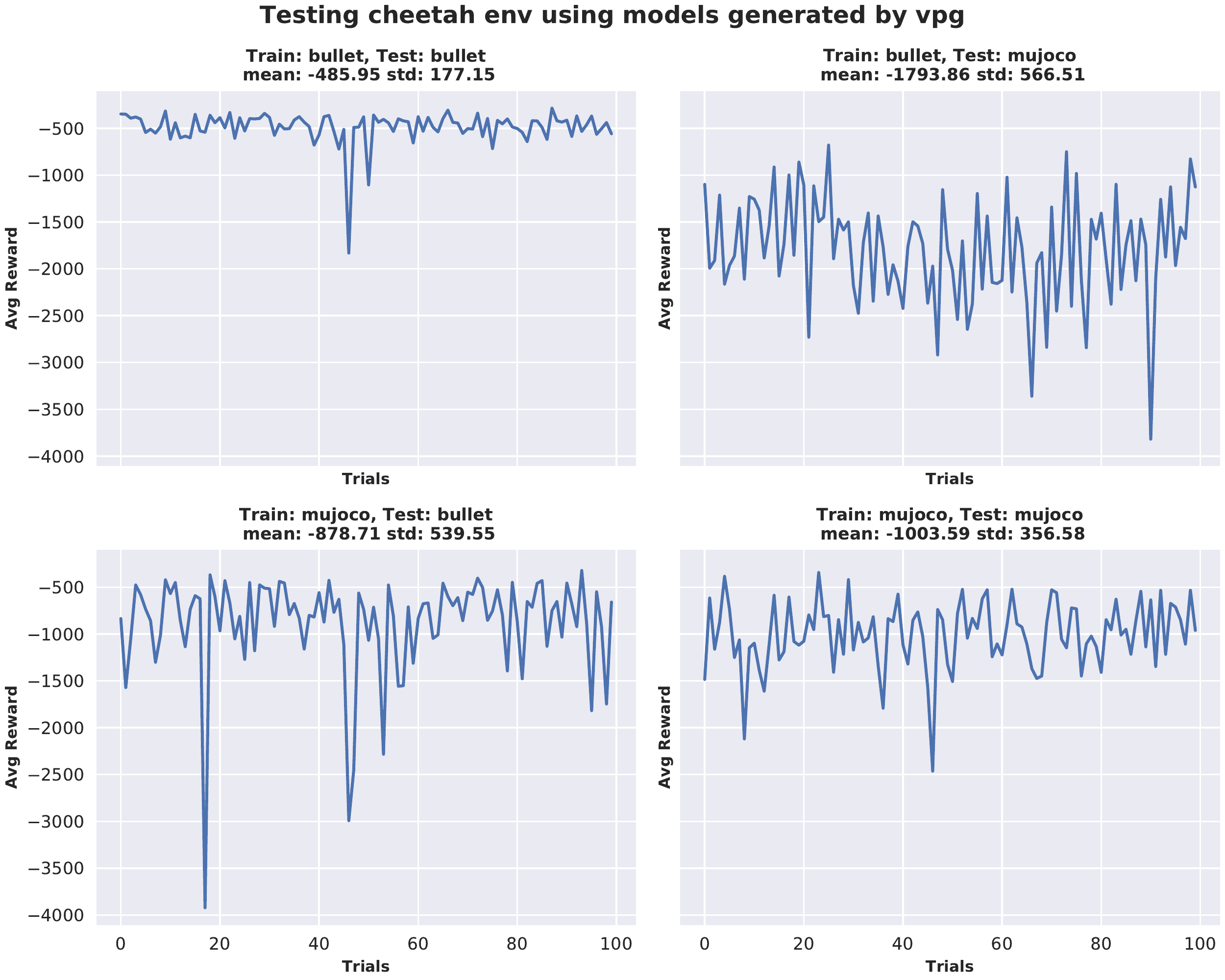}
		\caption{Comparison of generalization of VPG on half cheetah task across physics engines}
		\label{fig:cheetah vpg test}
	\end{figure}

\subsection{Ant}
	
	\begin{figure}[!h]
     \centering
     \begin{subfigure}[b]{0.6\textwidth}
         \centering
         \includegraphics[width=\textwidth]{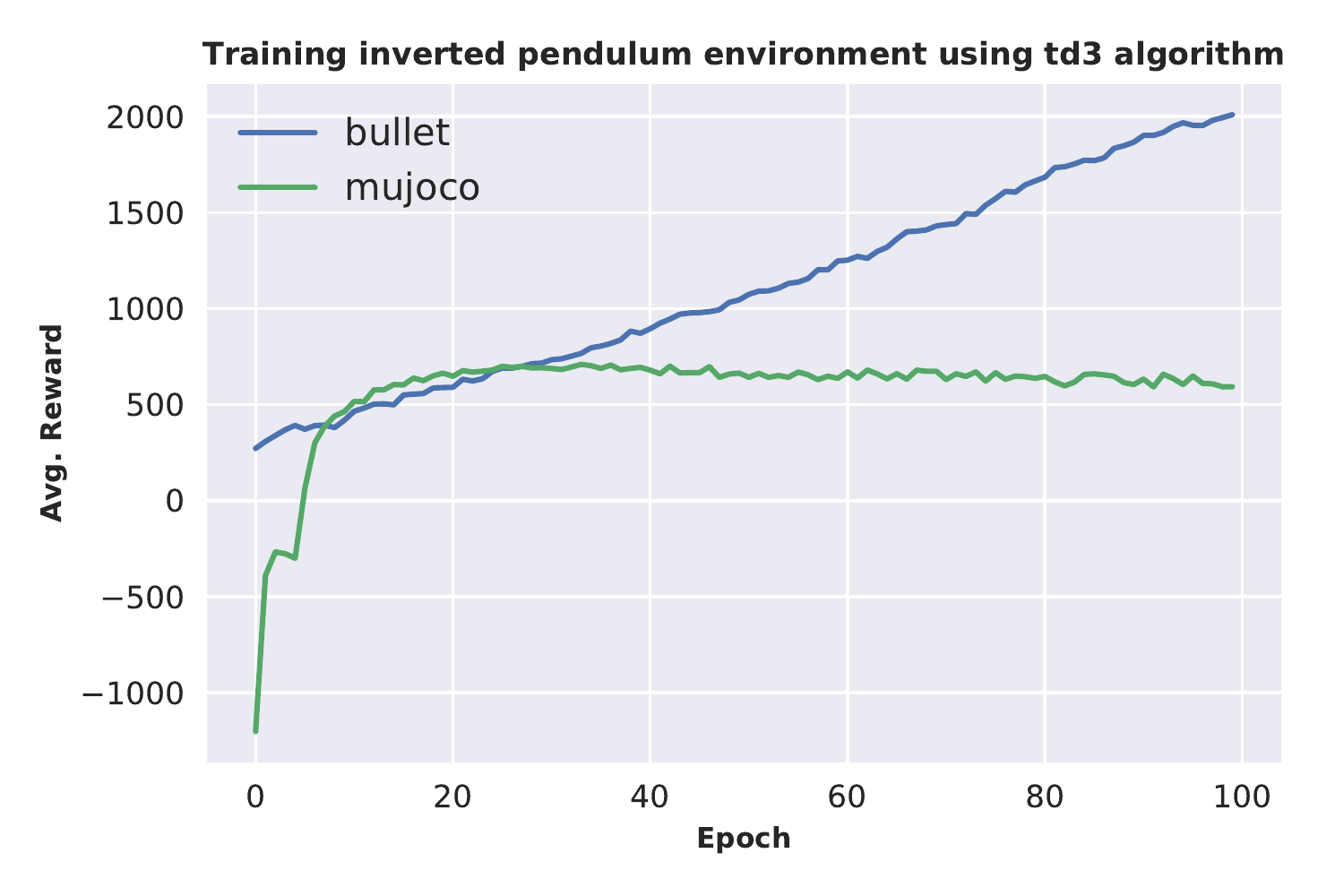}
         \caption{}
         \label{fig:ant td3 train}
     \end{subfigure}
     \hfill
     \begin{subfigure}[b]{0.6\textwidth}
         \centering
         \includegraphics[width=\textwidth]{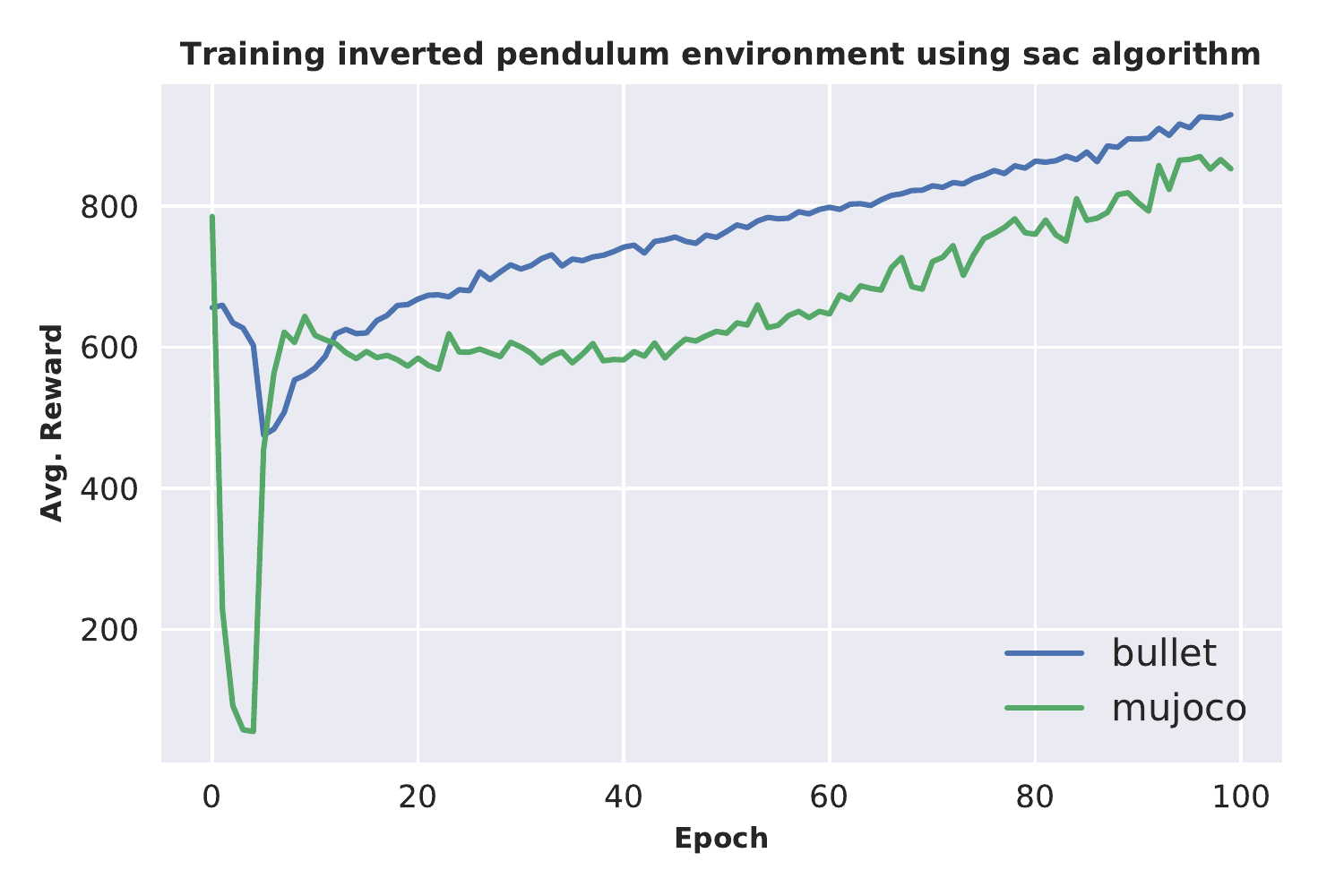}
         \caption{}
         \label{fig:ant sac train}
     \end{subfigure}
        \caption{Performance as a function of average reward over 100 different random seeds during the course of training. (a) Performance of TD3 on ant task across different physics engines. (b) Performance of SAC on ant task across different physics engines.}
        \label{fig:ant training}
	\end{figure}
	
	\begin{figure}[!h]
		\centering
		\includegraphics[width=0.75\textwidth]{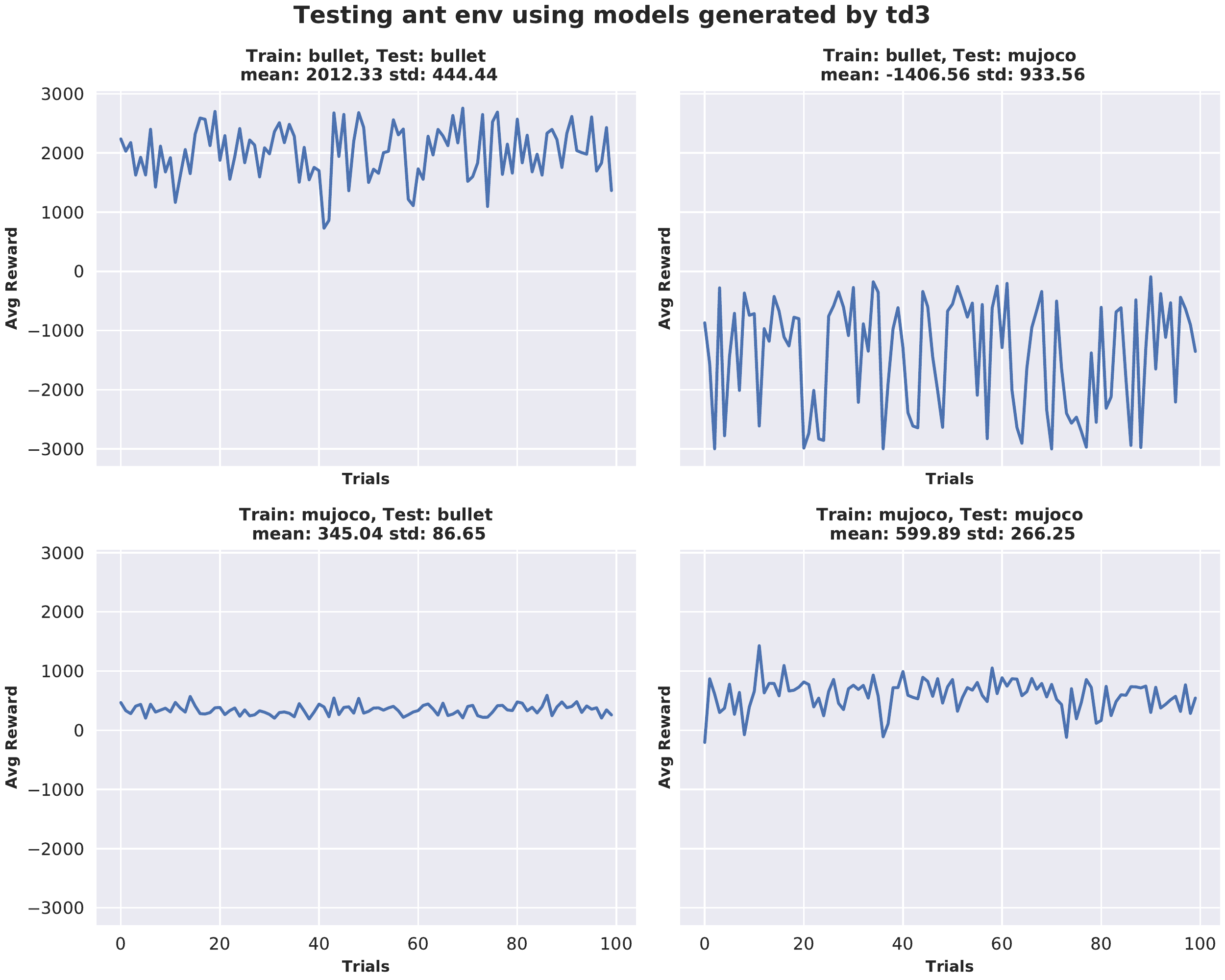}
		\caption{Comparison of generalization of TD3 on ant task across physics engines}
		\label{fig:ant td3 test}
	\end{figure}
		
	\begin{figure}[H]
		\centering
		\includegraphics[width=0.75\textwidth]{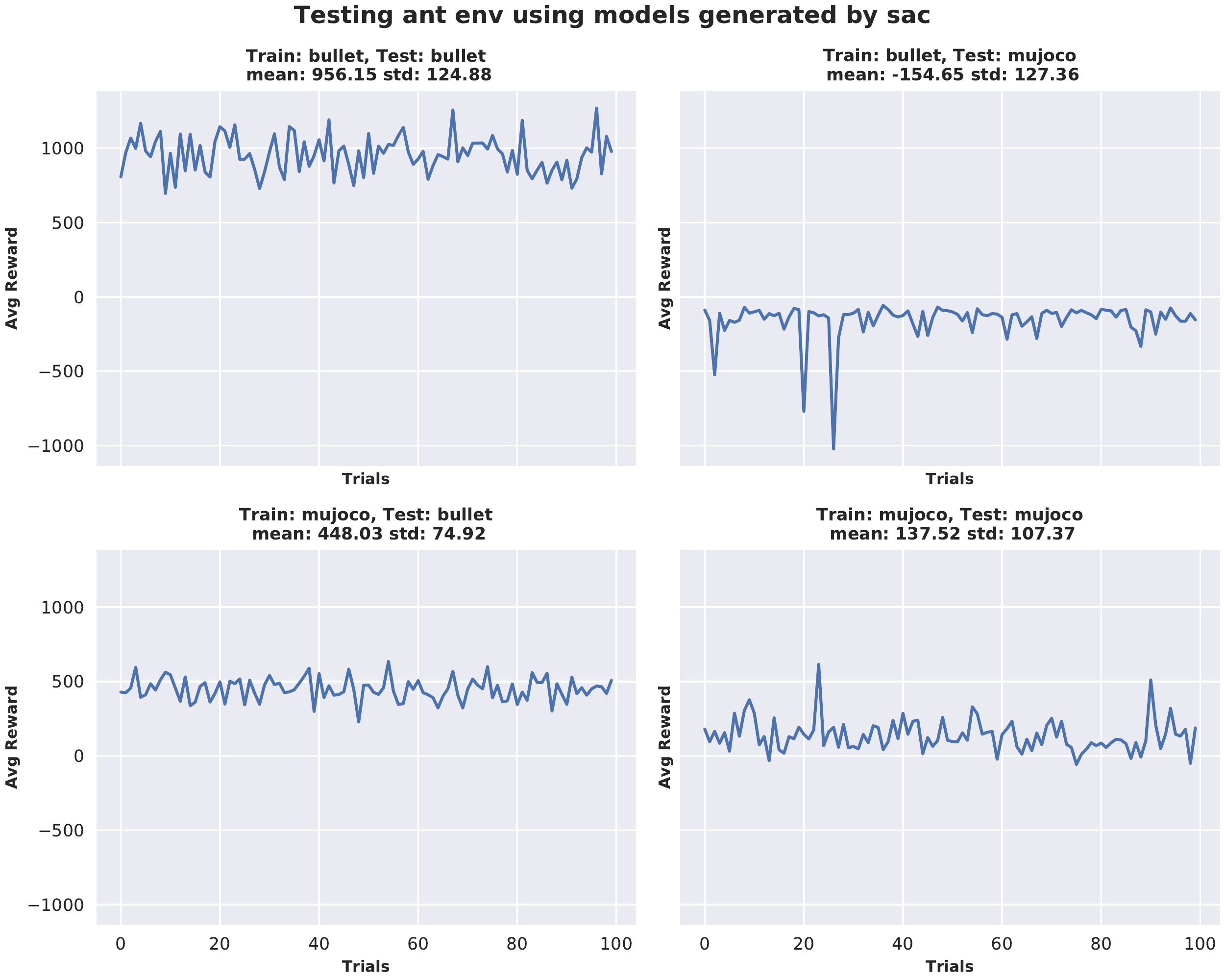}
		\caption{Comparison of generalization of SAC on ant task across physics engines}
		\label{fig:ant sac test}
	\end{figure}
\end{document}